\documentclass[journal]{IEEEtran}

\ifCLASSINFOpdf
\else
\fi

\usepackage{cite}
\usepackage{times}
\usepackage{algorithmic}
\usepackage{array}
\usepackage{mdwmath}
\usepackage{mdwtab}
\usepackage{eqparbox}
\usepackage{cite}
\usepackage{url}
\usepackage{multicol,multirow}
\usepackage[cmex10]{amsmath}
\usepackage{makeidx}
\usepackage{diagbox}
\usepackage{rotating,soul}
\usepackage{wrapfig}
\usepackage{booktabs}
\usepackage[usenames,dvipsnames]{color}
\usepackage{ragged2e}
\usepackage{caption}
\usepackage{pifont}
\usepackage{overpic}
\makeindex

\usepackage{geometry}
\RequirePackage{silence}
\DeclareGraphicsExtensions{.pdf,.jpg,.png}

\usepackage[colorlinks,bookmarksnumbered,bookmarksopen,linkcolor=black,citecolor=black,urlcolor=black]{hyperref}

\newcommand{\figref}[1]{Fig.~\ref{#1}}
\newcommand{\tabref}[1]{Tab.~\ref{#1}}
\newcommand{\eqnref}[1]{Eqn.~(\ref{#1})}
\newcommand{\secref}[1]{Sec.~\ref{#1}}

\newcommand{\myPara}[1]{\vspace{.05in}\noindent\textbf{#1.}\quad}

\def\ie{\emph{i.e.~}}
\def\eg{\emph{e.g.~}}

\def\etc{{\em etc.~}}

\newcommand{\sArt}{state-of-the-art~}
\newcommand{\addFig}[1]{}
\newcommand{\addFigs}[1]{}

\graphicspath{{./figs/}}

\newlength\savedwidth
\newcommand{\whline}[1]{\noalign{\global\savedwidth\arrayrulewidth \global\arrayrulewidth #1}%
                   \hline \noalign{\global\arrayrulewidth\savedwidth}}

\begin{document}

\title{Dynamic Feature Integration for Simultaneous 
Detection of Salient Object, Edge and Skeleton}

\author{Jiang-Jiang Liu, Qibin Hou, and Ming-Ming Cheng,~\IEEEmembership{Senior~Member,~IEEE}%
\thanks{J.J. Liu, and M.M. Cheng are with
      College of Computer Science, Nankai University.
      M.M. Cheng is the corresponding author (cmm@nankai.edu.cn).}
\thanks{Q. Hou is with National University of Singapore.}
}
\markboth{~}%
{Liu \MakeLowercase{\textit{et al.}}: Dynamic Feature Integration for Simultaneous Detection of Salient Object, Edge and Skeleton}

\maketitle

\begin{abstract}
%
In this paper, we solve three low-level pixel-wise vision problems, including
salient object segmentation, edge detection, and skeleton extraction, 
within a unified framework.
We first show some similarities shared by these tasks and then demonstrate
how they can be leveraged for developing a unified framework that can be
trained end-to-end.
In particular, we introduce a selective integration module that
allows each task to dynamically choose features at different levels
from the shared backbone based on its own characteristics.
Furthermore, we design a task-adaptive attention module, aiming at
intelligently allocating information for different tasks according to 
the image content priors.
%
%
To evaluate the performance of our proposed network on these tasks, 
we conduct exhaustive experiments on multiple representative datasets.
We will show that though these tasks are naturally quite different, 
our network can work well on all of them and even perform better than 
current single-purpose state-of-the-art methods.
In addition, we also conduct adequate ablation analyses that provide 
a full understanding of the design principles of the proposed framework.
To facilitate future research, source code will be released.
%

\end{abstract}


\IEEEpeerreviewmaketitle

\section{Introduction} \label{sec:introduction}
\IEEEPARstart{W}{ith} the rapid popularization of mobile devices, 
more and more deep learning based computer vision applications have been being ported 
from computer platforms to mobile platforms.
%
Many low-level computer vision tasks, benefiting from their 
category-agnostic characters,
act as fundamental components in mobile devices.
For example,
when using a smartphone to photograph, 
many supporting tasks are running in the background 
to assist users with better pictures and provide real-time effect previews.
Single-camera smartphones usually apply the salient object segmentation task
to simulate the bokeh effect that requires depth information \cite{hou2016deeply}.
To help users taking pictures with more visual pleasing compositions,
the edge detection task is adopted to
obtain structure information \cite{RcfEdgePami2019}.
And the skeleton extraction task plays an important role in
supporting taking photos by gesturing and 
instructing users with more interesting poses \cite{zhao2018hifi}.
However, due to the limited storage and computing resources of mobile devices,
it is inconvenient and inefficient to store the pre-trained models
for every different applications and perform multiple different tasks sequentially.

One feasible solution is to perform the aforementioned tasks within 
a single model 
but there exist two main challenges.
One is how to learn different tasks simultaneously while the other is 
how to settle the divergence of feature domains and 
optimization targets of different tasks.
Most previous work \cite{kokkinos2017ubernet,wu2019mutual,liu2019simple,wu2019stacked} 
solved the first challenge by 
observing the characteristics owning by different tasks and 
designing specialized network structures for each task manually.
They assumed that all the tasks learned jointly 
are complementary and some tasks are auxiliary
(\eg utilizing extra edge information to help the salient object detection task 
with more accurate segmentations in edge areas).
Usually the performances of the auxiliary tasks are sacrificed and ignored.
But when facing the second challenge that the tasks being solved 
are contrasting, as demonstrated in \figref{fig:teaser},
directly applying these methods often fails.
As shown in the 3rd row of \tabref{tab:ablation_results}, 
when trained jointly with the other two tasks,
the performance of skeleton extraction is badly damaged.

The design criterion of previous work is usually task-oriented and specific, 
greatly restricting their applicability to other tasks \cite{kokkinos2017ubernet}.
From the standpoint of network architecture, in spite of 
three different tasks, all of them require multi-level features, 
though in varying degrees.
Salient object segmentation 
requires the ability to extract homogeneous regions and hence relies
more on high-level features \cite{hou2016deeply}.
Edge detection aims at detecting accurate boundaries and hence
needs more low-level features to sharpen the coarse edge maps 
produced by deeper layers \cite{xie2015holistically,maninis2017convolutional}.
Skeleton extraction \cite{shen2016object,ke2017srn} prefers 
a proper combination of low-, mid- and high-level information 
to detect scale-variant (either thick or thin) skeletons.
%
%
%
%
Thus, a natural question is whether it is possible to 
design an architecture that can coalesce these three contrasting low-level vision tasks 
into a unified but end-to-end trainable network with
no loss on the performance of each task.
\renewcommand{\addFig}[1]{\includegraphics[width=0.235\linewidth]{teaser/#1}}
\begin{figure}[tp]
  \centering
  \small
  \renewcommand{\tabcolsep}{0.3mm}
  \begin{tabular}{cccc}
    \addFig{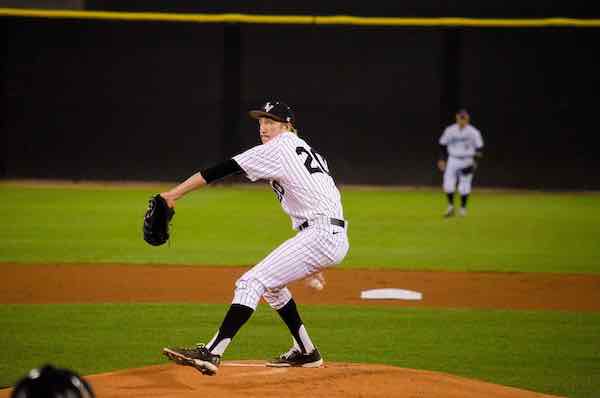} & \addFig{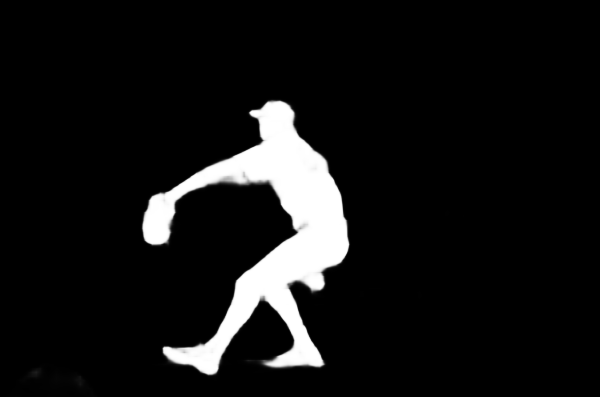} &
    \addFig{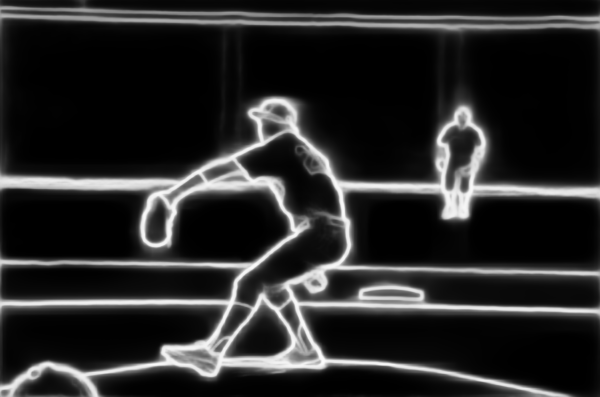} & \addFig{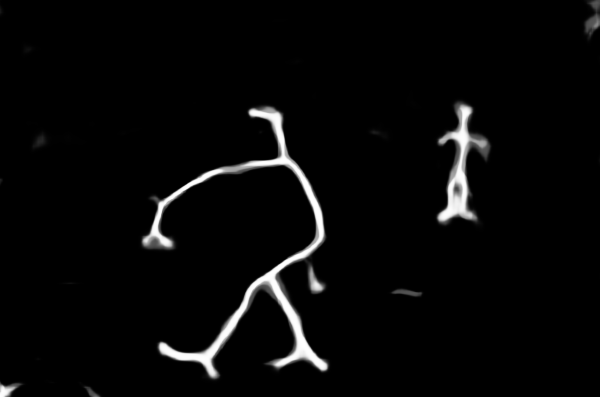} \\
     Image & Saliency & Edge & Skeleton 
  \end{tabular}
  \vspace{-3pt}
  \caption{An example case where information may conflict when
    learning saliency, edge, and skeleton simultaneously.
  The man in the behind is not salient yet has skeletons,
  And for edge detection, 
  it needs to detect all possible edge areas, whether being 
  salient or belonging to skeletons.
  All the predictions are performed by our approach.}
  \label{fig:teaser}
  \vspace{-5pt}
\end{figure}

\begin{figure*}[tp]
  \centering
  \includegraphics[width=1.0\linewidth]{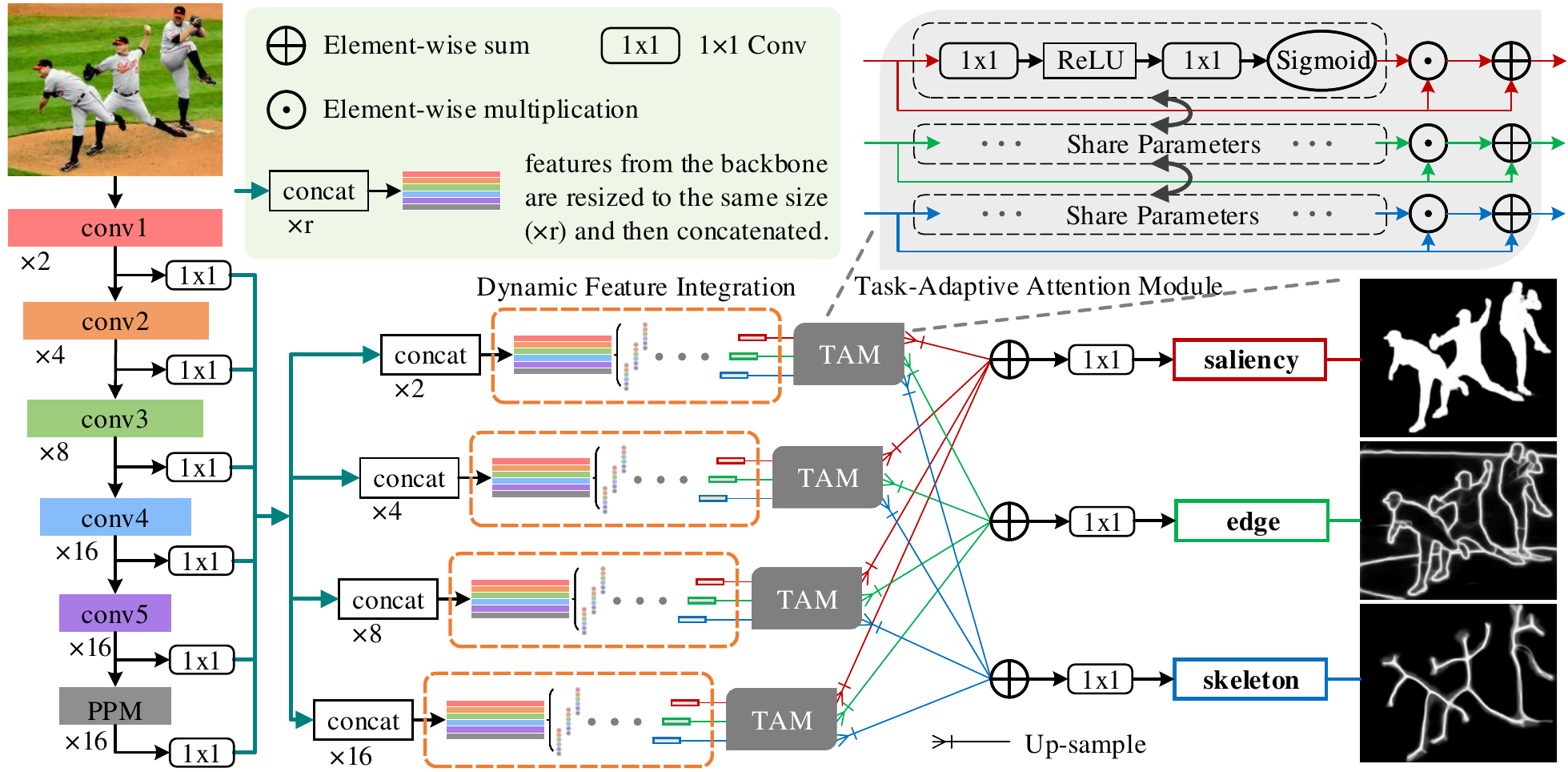}
  \caption{The overall pipeline of our proposed approach (best viewed in color).
  }\label{fig:arch}
  \vspace{-8pt}
\end{figure*}

Taking into account the different characteristics of each task, 
we present a novel, unified framework to settle the above challenges.
Specifically, our network comprises a shared backbone and
three task branches of identical design, as shown in \figref{fig:arch}.
To facilitate each task branch to select
appropriate features at different levels of the backbone automatically,
we introduce a dynamic feature integration strategy that is able to
choose favored features dynamically in an end-to-end learning manner.
%
This dynamic strategy can largely ease the process of architecture building
and promote the backbone to adjust its parameters for solving
multiple problems adaptively.
Then a task-adaptive attention module is adopted to
enforce the interchange of information among different task branches
in a separate-gather way.
By coupling previously independent branches, we can avoid 
the network to optimize asymmetrically.
Our approach is easy to follow
and can be trained end-to-end on a single GPU.
Without sacrifice on performance, it reaches a speed of 40 FPS 
performing the three tasks
simultaneously
when processing a $300\times400$ image.

To evaluate the performance of the proposed architecture, 
we compare it with the state-of-the-art methods of
the three tasks.
Experimental results show that our approach outperforms existing single-purpose methods 
on multiple widely used benchmarks.
Specifically, for salient object segmentation, 
compared to previous state-of-the-art works,
our method has a performance gain of 1.2\% in 
terms of F-measure
on average, over six popular datasets.
For skeleton extraction, we also improve the state-of-the-art results 
by 1.9\% in terms of F-measure on the SK-LARGE dataset \cite{shen2017deepskeleton}.
Furthermore, to let readers better understand the proposed approach, 
we conduct extensive ablation experiments on different components of the proposed architecture.

To sum up, the contributions of this paper can be summarized as follows:
(i) We design a dynamic feature integration
strategy to explore the feature combinations automatically
according to each input and task, 
and solve three contrasting tasks 
simultaneously in an end-to-end unified framework, running at 40 FPS;
(ii) We compare our multi-task approach 
with the single-purpose state-of-the-arts of each task and 
obtain better performances.

\section{Related Work} \label{sec:relatedWorks}


\subsection{Relevant Binary Tasks}\label{sec:relatedWorksBPT}

%

For salient object segmentation, 
traditional methods are mostly based on hand-crafted features \cite{cheng2015global,huang2017300,liu2011learning,jiang2013salient,perazzi2012saliency}.
With the popularity of CNNs, many methods
\cite{li2015visual,wang2015deep,zhao2015saliency,lee2016deep,SuperCNN_IJCV2015} started to use CNNs to extract features.
%
%
Some of them \cite{wangsaliency,liu2016dhsnet,wang2018detect,zhang2018progressive,wang2019iterative,xu2019deepcrf} incorporated the idea of
iterative and recurrent learning to refine the predictions.
There are also works solving the problem from the aspect of fusing richer features \cite{li2016deep,li2017instance,luo2017non,zhang2017amulet,hou2016deeply,zhang2017learning,zhang2018bi,wu2019cascaded,xu2019structured,zeng2019towards}, introducing attention mechanism \cite{chen2018reverse,zhang2018progressive,liu2018picanet,zhao2019pyramid},
using multiple stages to learn the prediction in a stage-wise manner
\cite{xiao2018deep,wang2017stagewise,wang2018detect}, or
adding more supervisions to get predictions 
with sharper edges \cite{li2018contour,liu2019simple,wang2019salient,feng2019attentive,fan2018SOC,zhang2019capsal,qin2019basnet,wu2019mutual,wu2019stacked,su2019selectivity,zhao2019egnet,zhao2019optimizing,liu2019employing}.
%
%
For edge detection, early works \cite{canny1986computational,marr1980theory,torre1986edge} 
mostly relied on various gradient operators.
Later works \cite{konishi2003statistical,martin2004learning,arbelaez2011contour} further employed manually-designed features.
Recently, CNN-based methods solved this problem by using fully-convolutional networks
in a patch-wise \cite{ganin2014n,shen2015deepcontour,bertasius2015deepedge,hwang2015pixel} 
or pixel-wise prediction manner
\cite{xie2015holistically,kokkinos2015pushing,yang2016object,liu2016learning,maninis2017convolutional,wang2017deep,RcfEdgePami2019,he2019bi}.
For skeleton extraction, earlier methods \cite{yu2004segmentation,jang2001pseudo,majer2004influence} 
mainly relied on gradient intensity maps of natural images to extract skeletons.
Later, learning-based methods \cite{tsogkas2012learning,sironi2014multiscale,levinshtein2013multiscale,widynski2014local} 
viewed skeleton extraction as a per-pixel classification problem or a super-pixel clustering problem.
Recent methods \cite{shen2016object,ke2017srn,zhao2018hifi,wang2019deepflux}
designed powerful network structures 
considering this problem hierarchically.
%
Different from all the above approaches, our approach simultaneously
solves the three tasks within a unified framework instead of
learning each task with an individual network.

\subsection{Multi-Task Learning}\label{sec:relatedWorksMTL}

Multi-task learning (MTL) has a long history in the area of machine learning \cite{caruana1997multitask,doersch2017multi,evgeniou2004regularized,kumar2012learning}. 
%
Recently, many CNN-based MTL
methods had been proposed, most of which focused on the design of 
network architecture \cite{misra2016cross,kokkinos2017ubernet,ahn2019deep,strezoski2019many}, 
or loss functions to balance the importance of different tasks \cite{kendall2018multi,chen2018gradnorm}, or both of them \cite{liu2019end}.
%
%
Different work also solved different task combinations,
including: image classification in multiple domains \cite{rebuffi2017learning}; object recognition, localization, and detection \cite{sermanet2013overfeat,ren2015faster,he2017mask};
pose estimation and action recognition \cite{gkioxari2014r,kendall2015posenet,du2019cross};
semantic classes, surface normals, and depth prediction \cite{eigen2015predicting,teichmann2018multinet,misra2016cross,liu2019end,kendall2018multi,gao2019nddr}.
%
However, the majority of these methods focused on specific related tasks requiring datasets with
different types of annotations supported simultaneously.
Different from the above methods, we aim to incorporate the 
idea of dynamic feature integration into architecture design.
%
This allows our approach to learn multiple tasks together
based on training data from multiple individual datasets.
Moreover, unlike the previous methods \cite{kokkinos2017ubernet,wu2019mutual} 
which fix the strategies of how features integrate into network structures,
our approach can adjust the 
network connections to select features dynamically to facilitate multi-task training.
%



\section{Method} \label{sec:method}

%
In this section, instead of attempting to manually 
design an architecture that might work for all the three
tasks, we propose to encourage the network to 
dynamically select features at different levels
according to the favors of each task and the content of each input as described in 
\secref{sec:introduction}.
%
%
%

%

\begin{figure}[tp]
  \centering
  \includegraphics[width=1.\linewidth]{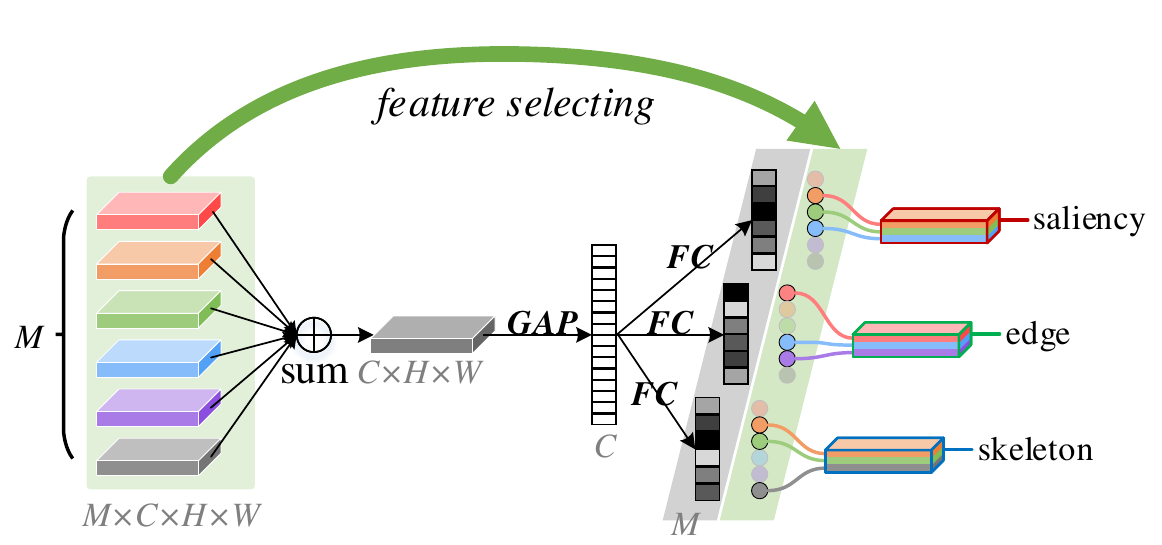}
  \caption{Detailed illustration of DFIM.
   It takes the set of features extracted from the backbone as input, 
   which is resized to the same size first.
   Then different stages of these features are dynamically selected for each task.
  }\label{fig:dfim}
\end{figure}

\subsection{Overall Pipeline}

We include three different tasks on multiple individual datasets (\ie, 
DUTS \cite{wang2017learning} for saliency, BSDS 500 \cite{arbelaez2011contour} and 
VOC Context \cite{mottaghi2014role} for edge,
SK-LARGE \cite{shen2016object} or SYM-PASCAL \cite{ke2017srn} for skeleton) 
within a unified network which can be trained end-to-end.
All the datasets are directly used following the existing single-purpose methods
proposed for each task with no extra processing.

The overall pipeline of the proposed framework is
illustrated in \figref{fig:arch}.
We employ the ResNet-50 \cite{He2016} network as the feature extractor.
We take the feature maps outputted by \texttt{conv\_1} as $S_1$, 
and the outputs by \texttt{conv2\_3}, \texttt{conv3\_4}, 
\texttt{conv4\_6}, and \texttt{conv5\_3} as $S_2$ to $S_5$, respectively.
We set the dilation rates of the $3\times3$ convolutional layers
in \texttt{conv5} to 2 as done in pixel-wise prediction tasks.
%
Moreover, we add a pyramid pooling module (PPM) \cite{zhao2016pyramid}
on the top of ResNet-50 to capture more global information as done
in \cite{wang2017stagewise,liu2019simple}.
The output is denoted as $S_6$.
%

%
Rather than manually fixing the feature integration strategy in the 
network structure as done 
in most of the previous single-purpose methods,
%
a serious of dynamic feature integration modules (DFIMs) 
of various output down-sampling rates 
(orange dashed rounded rectangles in \figref{fig:arch})
are arranged to integrate the features extracted from the backbone 
(\ie $\{S_i\}~(1\le i \le M, M=6)$)
dynamically and separately for the three tasks.
%
%
%
%
%
%
%
%

A task-adaptive attention module (TAM) is then followed
after each DFIM to intelligently allocate information across tasks,
preventing the network from tendentious optimization directions.
%
%
%
Finally, the corresponding feature maps outputted by the TAMs for each task 
are up-sampled and summarized and then followed by a $1\times1$ convolutional layer for 
final prediction, respectively.
%
%

\begin{figure}[tp]
  \centering
  \includegraphics[width=1.\linewidth]{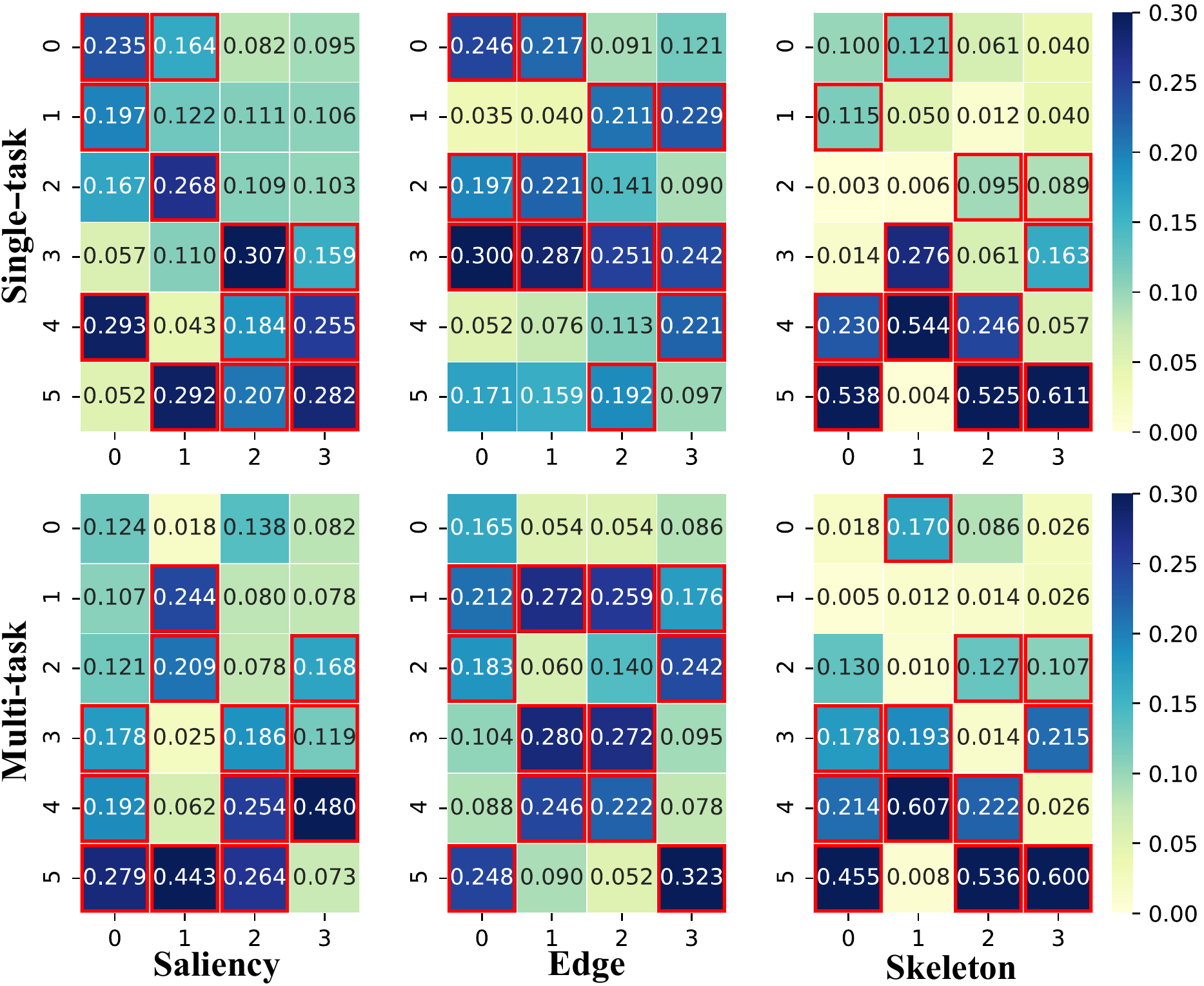}
  \vspace{-16pt}
  \caption{The weights of each stage of features been selected
  by each DFIM. 
  In each subplot, row indicates indexes of DFIMs, and column
  indicates stages of features.
  Only the top half stages (red rectangles) in each DFIM are kept.
  }\label{fig:dfim_softmax}
\end{figure}

\subsection{Dynamic Feature Integration}
It has been mentioned in many previous multi-task methods \cite{kokkinos2017ubernet,misra2016cross,kendall2018multi,liu2019end} that
the features required by different tasks vary greatly.
And most of them require multiple kinds of annotations within a single dataset,
which is difficult to obtain.
Differently, we utilize training data of different tasks from multiple individual datasets,
and it is more likely to meet circumstances where features required by different tasks
conflict, as demonstrated in \figref{fig:teaser}.
%
%
%
%
To solve this problem, we propose DFIM, which adjusts
the feature integration strategy dynamically according
to each task and input during both training and testing.
Compared to existing methods that integrate specific levels 
of features from the backbone based on manual observations 
of different tasks' characteristics,
DFIM learns the feature integration strategy.
%

%
%
%
%

\begin{figure*}[tp]
  \centering
  \includegraphics[width=1.\linewidth]{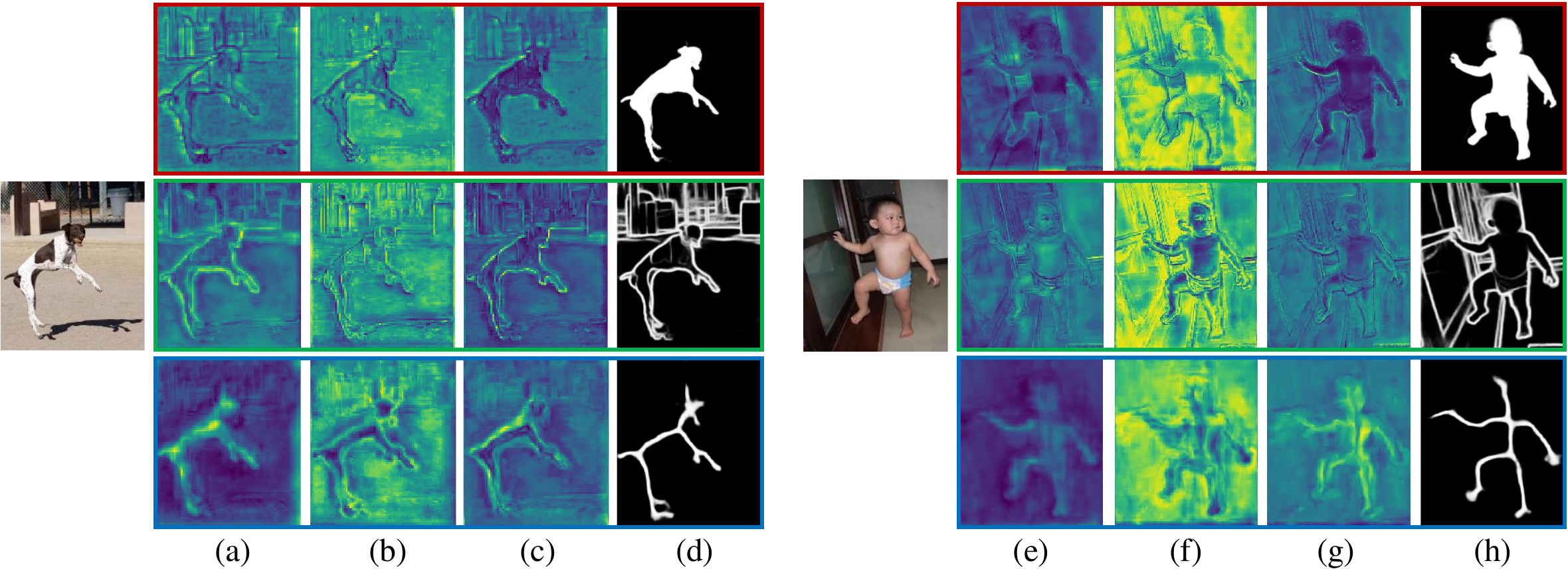}
  \vspace{-15pt}
  \caption{Visualizing feature maps around TAM. (a,e) Before TAM; (b,f) Attention maps in TAM;
  (c,g) After TAM; (d,h) Predictions. As can be seen, TAM can tailor the feature maps for each task adaptively
  and accordingly. From top to bottom: more distinct saliency parts, sharper edges, and
  stronger skeletons.
  }\label{fig:tam_vis}
\end{figure*}

To be specific, 
we take the set of features $\{S_i\}~(1\le i \le M, M=6)$ extracted from
the backbone as input for each DFIM.
%
And the demanded output down-sampling rate $\times r$ of each DFIM is 
determined during the network definition period. 
%
%
As illustrated in \figref{fig:dfim},
%
we first transfer $\{S_i\}~(1\le i \le M, M=6)$ to the 
same number of channels (\ie, $C \times$) and down-sampling rate (\ie, $\times r$),
denoted as $\{S_i^r\}~(1\le i \le M, M=6)$, 
through a $1\times1$ convolutional layer and bilinear interpolation, respectively.
To give DFIM a view of all the features to be selected,
we summarize $\{S_i^r\}$,
and follow it with a global average pooling (GAP) layer to create a compact and global
feature ($C\times$) as done in \cite{hu2018squeeze}.
For each $task \in \{saliency, edge, skeleton\}$, we use an independent fully connected (FC) layer to map
the $C\times$ feature to $M\times$ channels,
and then apply a softmax operator to transform the 
$M\times$ feature into the form of probability $\{p_i^{r, task}\}~(1\le i \le M)$
that could be used as an indicator to
select features.
Different from those who keep dense connections of $\{S_i^r\}$,
we only keep half the connections as $\{S_i^{r, task}\}~(1\le i \le M)$:
\begin{equation}
  S_i^{r, task} = 
  \begin{cases}
    p_i^{r, task} * S_i^r,  & \text{if $p_i^{r, task} \ge median(\{p_i^{r, task}\})$} \\
    0, & \text{else,}
    \end{cases} \label{eqn:dfi}
\end{equation}
in which $median(\cdot)$ means taking the median.
Thus the output of DFIM with down-sampling rate $\times r$ for the $task$ can be obtained with
\begin{equation}
D^{r, task} = \sum_i S_i^{r, task}. \label{eqn:dfi_sum}
\end{equation} 
%

%
By arranging a series of DFIMs of various down-sampling rates, 
we can obtain the dynamically integrated 
feature maps $\{D^{r, task}\}(r \in \{2,4,8,16\}, task \in \{saliency, edge, skeleton\})$,  
as shown in \figref{fig:arch}.
Since the feature integration strategies depend only on the input and task type,
the network is able to learn integration strategies for each input and 
task within a broader and more flexible feature combination space in an end-to-end manner.

\subsection{Task-Adaptive Attention}

%
As we utilize training data from multiple individual datasets,
the domain shifting \cite{caruana1997multitask,pan2009survey}
problem can not be ignored.
%
%
How to coalesce the information from diverse
datasets effectively and efficiently is indispensable to the maintaining
of the overall performance across all tasks.
%
%
As illustrated in the first row (Single-task) of \figref{fig:dfim_softmax}, 
the levels of features that different tasks favor vary greatly.
%
%
If we use the task-specific feature maps $\{D^{r, task}\}(r \in \{2,4,8,16\})$ 
generated by DFIMs
for the prediction of each task directly,
the gradients of some task to the shared parts of the network 
may bias distinctly from the other tasks, hence deflecting the optimization
direction to local minimums and causing under-fitting. 

To this end, we propose to let the network have the ability
of intelligently allocating information for different tasks after
the shared features from the backbone are dynamically integrated and 
tailored for each task.
As shown in the top right corner of \figref{fig:arch}, 
the output feature maps $\{D^{r, task}\}(task \in \{saliency, edge, skeleton\})$ from the DFIM 
with a down-sampling rate of $\times r$ 
are further rescaled by being forwarded to a TAM.
The parameters in TAM are shared across tasks for the exchange of information.
Compared to using the outputs for each task directly,
the additional modeling of all tasks' relations 
gives DFIM the 
ability to adaptively adjust each task's influences
on the shared backbone 
by considering the content of input and all task's characteristics simultaneously.
%
%
%
%
%
%
TAM forces the interchange of information across tasks even after
the feature maps are separated and tailored for each task.
This is quite different from previous 
methods \cite{kokkinos2017ubernet,liu2019end,wu2019mutual},
which kept different tasks' branches independent of each other until the end.
%
%

To have a better perception,
we visualize the intermediate feature maps around TAM in \figref{fig:tam_vis}.
%
As can be seen, 
in the 1st row, for salient object segmentation,
before TAM (a,e), it is hard to distinguish the dog (child) from the background.
The attention maps learned in TAM (b,f) erase the activation of background
effectively.
And after TAM (c,g), the dog (child) is highlighted clearly.
In the 2nd row, for edge detection,
the feature maps after TAM (c,g) have obvious
thinner and sharper activations in the areas where
edges may locate compared to the thick and blur activations before TAM (a,e).
A similar phenomenon can also be observed for the skeleton extraction task.
As shown in the last row, 
the skeletons of the dog (child) become stronger and clearer after TAM.
%
All the aforementioned discussions verify the significant effect of TAM on better
allocating the information for different tasks.

\section{Experiment Setup} \label{sec:experiment_setup}

In this section, we describe the experiment setups, including
the implementation details of the proposed network, 
the used datasets, the training procedure, and the evaluation metrics for the three tasks.
%
%


\myPara{Implementation Details}
We implement the proposed method based on PyTorch\footnote{\url{https://pytorch.org}}.
All experiments are carried out on a workstation with an Intel
Xeon 12-core CPU (3.6GHz), 64GB RAM, and a single NVIDIA RTX-2080Ti GPU.
We use the Adam \cite{kingma2014adam} optimizer 
with an initial learning rate of 5e-5 and a weight decay of 5e-4.
Our network is trained for 12 epochs in total, and the 
leaning rate is divided by 10 after 9 epochs.
The parameters of the backbone (\ie, ResNet-50 \cite{He2016}) of
our network are initialized with ImageNet \cite{krizhevsky2012imagenet} pre-trained model,
while all other parameters are randomly initialized.
Group normalization \cite{wu2018group} is applied after each convolutional layer 
except for the backbone.
The optimization configurations for all parameters in our network
are identical, except for the parameters of the  
batch normalization layers of the backbone
are frozen during both training and testing.

\myPara{Datasets}
We use individual datasets for different tasks, and 
each dataset only has one kind of annotation.
The detailed configurations are listed in \tabref{tab:datasets}.
And all the datasets are directly used following the existing single-purpose methods
proposed for each task \cite{zhang2017amulet,RcfEdgePami2019,zhao2018hifi} with no extra pre-processing.

\begin{table}[tp]
  \centering
  \scriptsize
  \setlength\tabcolsep{0.4mm}
  \renewcommand{\arraystretch}{1.3}
  \caption{The datasets we use for
  training and testing.}\label{tab:datasets}
  \begin{tabular}{l|c|c|c|c}
  \whline{1pt}
     Task & Training & \#Images & Testing & \#Images \\ \whline{1pt}
     \multirow{3}{*}{Saliency} & \multirow{3}{*}{DUTS-TR \cite{wang2017learning}} & \multirow{3}{*}{10553} & ECSSD \cite{yan2013hierarchical}, 
    PASCAL-S \cite{li2014secrets}, & 1000, 850, \\ 
    & & & DUT-OMRON \cite{yang2013saliency}, SOD \cite{movahedi2010design}, & 5166, 300, \\
    & & & HKU-IS \cite{li2015visual}, DUTS-TE \cite{wang2017learning} & 1447, 5019 \\ \whline{0.8pt}
    \multirow{2}{*}{Edge} & BSDS500 \cite{arbelaez2011contour} \& & 300 +  & \multirow{2}{*}{BSDS500 \cite{arbelaez2011contour}} & \multirow{2}{*}{200} \\
                          & VOC Context \cite{mottaghi2014role} & 10103 & & \\ \whline{0.8pt}
    \multirow{2}{*}{Skeleton} & SK-LARGE \cite{shen2017deepskeleton} & 746 & SK-LARGE \cite{shen2017deepskeleton} & 745 \\ \cline{2-5}
     & SYM-PASCAL \cite{ke2017srn} & 648 & SYM-PASCAL \cite{ke2017srn} & 788 \\ 
  \whline{1pt}
  \end{tabular}
\end{table}

\myPara{Training Procedure}
To jointly solve three different tasks from three individual datasets in an end-to-end way,
for each iteration, 
we randomly sample an image-groundtruth pair for each of the three tasks, respectively.
Then sequentially, each of the three image-groundtruth pairs is
forwarded to the network,
and the corresponding loss is calculated.
At last, we simply summarize the three calculated losses (integers), backward through
the network,
and then take an optimization step.
All the other training procedures are identical to typical single-purpose methods.

\myPara{Loss Functions}
We define the loss functions of the three tasks
as most of the previous single-purpose methods.
We use standard binary cross-entropy loss for salient object segmentation \cite{hou2016deeply,zhang2017amulet} and 
balanced binary cross-entropy loss \cite{xie2015holistically,RcfEdgePami2019,zhao2018hifi} 
for edge detection and skeleton extraction.
We simply summarize the losses of the three tasks as the overall loss.
The detailed formulas of the loss functions we used are as follows.
Given an image's prediction map $\hat{Y}$
and its corresponding groundtruth map $Y$, for all pixels $(i, j)$, 
we compute the standard binary cross-entropy loss as:
\begin{small}
\begin{equation}
  \begin{aligned}
  \mathcal{L}_{standard} (\hat{Y}, Y) =& - \sum_{i, j} [Y(i, j) \cdot \log \hat{Y}(i, j)\\
  &+ (1 - Y(i, j)) \cdot \log (1 - \hat{Y}(i, j))],
  \end{aligned}
\end{equation}
\end{small}
and the balanced binary cross-entropy loss as:
  \begin{small}
  \begin{equation}
    \begin{aligned}
    \mathcal{L}_{balanced} (\hat{Y}, Y) =& - \sum_{i, j} [ \beta \cdot Y(i, j) \cdot \log \hat{Y}(i, j)\\
    &+ (1 - \beta) \cdot (1 - Y(i, j)) \cdot \log (1 - \hat{Y}(i, j))],
    \end{aligned}
  \end{equation}
\end{small}
where $\beta = |Y^-| / |Y^+ + Y^-| $. 

\begin{itemize}
  \item
  Salient Object Segmentation: 
\begin{small}
\begin{equation}
  \mathcal{L}_{sal} (\hat{Y}_{sal}, Y_{sal}) = \mathcal{L}_{standard} (\hat{Y}_{sal}, Y_{sal}).
\end{equation}
\end{small}
\item
Edge Detection: 
\begin{small}
\begin{equation}
  \mathcal{L}_{edg} (\hat{Y}_{edg}, Y_{edg}) = \mathcal{L}_{balanced} (\hat{Y}_{edg}, Y_{edg}),
\end{equation}
\end{small}
in which $Y^+$ refers to the edge pixels and $Y^-$ refers to the non-edge pixels.
\item
Skeleton Extraction: 
\begin{small}
\begin{equation}
  \mathcal{L}_{ske} (\hat{Y}_{ske}, Y_{ske}) = \mathcal{L}_{balanced} (\hat{Y}_{ske}, Y_{ske}),
\end{equation}
\end{small}
in which $Y^+$ refers to the skeleton pixels and $Y^-$ refers to the non-skeleton pixels.
\end{itemize}

\begin{table}[tp]
  \centering
  \footnotesize
  \setlength\tabcolsep{1.3mm}
  \renewcommand{\arraystretch}{1.3}
  \caption{The composition of the proposed network's parameters. As can be seen, 
  the feature extractor (ResNet-50 \& PPM) and the shared parts take up the majority.
    }\label{tab:parameters}
  \begin{tabular}{c|c|c|c||c|c|c}
  \whline{1pt}
     \multicolumn{7}{c}{Total: 29.57M} \\ \hline
     \multicolumn{4}{c||}{Shared: 27.01M (91.34\%)} & \multicolumn{3}{c}{Specific: 2.56M (8.66\%)} \\ \whline{0.8pt}
     ResNet-50 & PPM & DFIMs & TAMs & Saliency & Edge & Skeleton \\ \hline
     23.46M & 1.31M & 1.42M & 0.83M & 0.85M & 0.85M & 0.85M \\
     79.34\% & 4.43\% & 4.80\% & 2.81\% & 2.87\% & 2.87\% & 2.87\% \\
  \whline{1pt}
  \end{tabular}
\end{table}

\myPara{Evaluation Criteria}
For salient object segmentation, we use F-measure score ($F_\beta$), 
mean absolute error (MAE),
precision-recall (PR) curves, and S-measure \cite{fan2017structure} for evaluation.
And the hyper-parameter $\beta^2$ in $F_\beta$ is set to 0.3 as done 
in previous work to weight precision more than recall.
For edge detection, we use the fixed contour threshold (ODS) and
per-image best threshold (OIS) as our measures. 
Before evaluation, 
we apply the standard non-maximal suppression (NMS) algorithm to get thinned edges.
For skeleton extraction, the skeleton maps are NMS-thinned before evaluation.
A series of precision/recall pairs are then obtained by applying different thresholds
to the thinned skeleton maps to draw the PR-curve.
And the F-measure score is obtained under the optimal threshold over the whole dataset.

\begin{table*}[tp!]
  \centering
  \footnotesize
  \renewcommand{\arraystretch}{1.2}
  \renewcommand{\tabcolsep}{1.2mm}
  \caption{
  Quantitative salient object segmentation, edge detection and 
  skeleton extraction results five widely used datasets.
  `Single-Task' means directly apply our method whilst only
  performing a single task.
  The best results in each column are highlighted in \textbf{bold}.
}\label{tab:ablation_results}
  \begin{tabular}{c|c|c||ccc|ccc|ccc||cc||c}
  \whline{1pt}
  \multirow{3}*{No.} & \multirow{3}*{DFIM} & \multirow{3}*{TAM} & \multicolumn{9}{c||}{Saliency} & \multicolumn{2}{c||}{Edge} & Skeleton \\
  \cline{4-15} 
  &  &  & \multicolumn{3}{c|}{PASCAL-S \cite{li2014secrets}} & \multicolumn{3}{c|}{DUT-OMRON \cite{yang2013saliency}} & \multicolumn{3}{c||}{DUTS-TE \cite{wang2017learning}} & \multicolumn{2}{c||}{BSDS 500 \cite{arbelaez2011contour}} & SK-LARGE \cite{shen2017deepskeleton}  \\
   \cline{4-15}
   &  &  & $F_\beta$~$\uparrow$ & MAE~$\downarrow$ & $S_m$~$\uparrow$ & $F_\beta$~$\uparrow$ & MAE~$\downarrow$ & $S_m$~$\uparrow$ & $F_\beta$~$\uparrow$ & MAE~$\downarrow$ & $S_m$~$\uparrow$ & ODS~$\uparrow$ & OIS~$\uparrow$ & $F_m$~$\uparrow$ \\
  \whline{1pt}
  \multicolumn{15}{l}{\textbf{Our Method (Single-Task)}} \\ \whline{1pt}
  1 & sparse & w/o & 0.860 & 0.075 & 0.849 & 0.811 & 0.059 & 0.835 & 0.875 & 0.042 & 0.878 & 0.815 & 0.831 & 0.749  \\ 
  2 & sparse & independent & 0.859 & 0.081 & 0.849 & 0.817 & 0.060 & 0.835 & 0.880 & 0.045 & 0.878 & 0.812 & 0.826 & 0.746  \\ \whline{1pt}
  \multicolumn{15}{l}{\textbf{Our Method (Multi-Task)}} \\ \whline{1pt}
  3 & identity & w/o & 0.877 & \textbf{0.062} & \textbf{0.865} & 0.818 & 0.056 & 0.836 & 0.885 & \textbf{0.038} & 0.886 & 0.811 & 0.828 & 0.708 \\ 
  4 & dense & w/o & 0.872 & 0.064 & 0.859 & 0.813 & 0.056 & 0.833 & 0.877 & 0.039 & 0.881 & 0.810 & 0.825 & 0.740 \\ 
  5 & sparse & w/o & 0.874 & 0.064 & 0.862 & 0.817 & 0.056 & \textbf{0.842} & 0.884 & \textbf{0.038} & \textbf{0.887} & 0.818 & 0.834 & 0.744  \\ \hline
  6 & sparse & independent & 0.873 & 0.065 & 0.861 & 0.815 & 0.057 & 0.836 & 0.879 & 0.039 & 0.883 & 0.815 & 0.832 & \textbf{0.753}  \\
  7 & sparse & share & \textbf{0.880} & 0.065 & \textbf{0.865} & \textbf{0.829} & \textbf{0.055} & 0.839 & \textbf{0.888} & \textbf{0.038} & \textbf{0.887} & \textbf{0.819} & \textbf{0.836} & 0.751  \\
  \whline{1pt}
  \multicolumn{15}{l}{\textbf{Other Methods (Multi-Task)}} \\ \whline{1pt}
  8 & \multicolumn{2}{c||}{$\text{UberNet}_\text{17}$ \cite{kokkinos2017ubernet}} & 0.823 & - & - & - & - & - & - & - & - & 0.785 & 0.805 & - \\ \hline
  9 & \multicolumn{2}{c||}{$\text{MLMS}_\text{19}$ \cite{wu2019mutual}} & 0.853 & 0.074 & 0.844 & 0.793 & 0.063 & 0.809 & 0.854 & 0.048 & 0.862 & 0.769 & 0.780 & -  \\
  \whline{1pt}
  \end{tabular}
\end{table*}

\section{Ablation Studies} \label{sec:ablation}
In this subsection, 
we first analyse the composition of parameters of the proposed 
model.
Then we investigate the effectiveness of the proposed DFIM 
by conducting experiments on both single- and multi-task settings.
Finally we show the effect of 
TAM with a better overall convergence and performance.

\subsection{Composition of Parameters}
We list the composition of the parameters of our network in \tabref{tab:parameters}.
As can be seen, 91.34\% of the parameters are shared across tasks 
where the feature extractor (ResNet-50 \& PPM) takes up 91.71\%.
And the shared parts of DFIMs and TAMs only bring in 2.25M (8.33\%) parameters.
Each task owns 0.85M (2.87\%) task-specific parameters, respectively.
The polarized composition of parameters proves the efficiency and effectiveness of the proposed
approach.
By taking advantage of the shared features extracted from the backbone
and adaptively recombining them, more parameters and space can be saved.
At the meanwhile, by handing the design of feature integration strategies 
to the network itself, less human interaction is required.

\subsection{Dynamic Feature Integration} \label{sec:ablation_dfi}

\myPara{Effectiveness of Dynamic Feature Integration}
As shown in the 1st row of \tabref{tab:ablation_results},  
directly applying our method whilst only performing a single task
can obtain comparable results with the \sArt methods on the salient object detection and 
edge detection tasks.
And a bigger promotion can be observed on the skeleton extraction task (1.7\%).
This indicates that the proposed DFIM is able to adjust it's feature selecting strategies
according to the characteristics of the target task being solved.
Rather than engineering specific network structure for different tasks manually as 
usually done in the previous methods, DFIM requires less human interactions.

When the three tasks are learned jointly (the 5th row of \tabref{tab:ablation_results}),
the performance of salient object segmentation task is 
promoted by a clear margin on three datasets in nearly all terms.
This is consistent with previous researches that edge information can help the salient
object segmentation task with more accurate segmentation in edge areas.
The performance of edge detection task also increases, indicating that 
the edges of salient objects may provide useful guidance signals as well.
The skeleton extraction task only drops slightly.

To have a numerical estimation of the difficulty in jointly training the three tasks,
we set up a baseline by
removing the dynamic feature selecting process in DFIM 
(the 3rd row in \tabref{tab:ablation_results}, marked as `identity').
This means that all the operations after the features extracted from
the backbone being summarized are removed, as shown in \figref{fig:dfim}.
This also equals to replacing \eqnref{eqn:dfi_sum} with 
\begin{equation}
  D^{r, task} = \sum_i S_i^{r}. \label{eqn:dfi_sum_identity}
\end{equation} 
By comparing the `identity' version with the proposed `sparse' feature selecting
version (the 5th row), 
we can observe clear drops on the tasks of edge detection and 
skeleton extraction, 0.7\% and 3.6\%, respectively.
These phenomenons demonstrate that
simply fusing all levels of features damages the detection of edge and 
skeleton.
It's difficult to design network structures manually when the involved tasks
have distinct optimization targets and take training samples from different datasets.
Similar circumstance occurs in previous works \cite{kokkinos2017ubernet,wu2019mutual}, 
where the performance of partial tasks decrease dramatically when solving different tasks jointly,
as shown in the last two rows of \tabref{tab:ablation_results},
But with DFIM, by letting the network itself to integrate features dynamically and accordingly, 
all of the three tasks perform comparably to training each task separately.

\myPara{Dynamically Learned Integration Strategies}
To have a better understanding of what feature integration strategies
have been learned by our proposed method,
we randomly select 100 images from each of the testing set of 
DUTS (saliency), BSDS500 (edge) and SK-LARGE (skeleton)
to form up a 300-image set.
By forwarding these images, 
we average the $\{p_i^{r, task}\}$ values of all images, which are the indicators for
features selecting.
We plot the probabilities of each stage of features from the backbone been
selected by each DFIM for different tasks in \figref{fig:dfim_softmax}.
%
If we compare the subplots column-wise,
the stages of features preferred by different tasks 
vary greatly.
This may explain why a good performing architecture for one task does not work
on the other tasks \cite{RcfEdgePami2019,hou2016deeply,wang2019deepflux}.
If we compare the subplots row-wise,
the stages of features been selected when each of the 
three tasks is separately trained in a single-task manner also differ
greatly from those when they are jointly trained in a multi-task manner.
This may be the reason why each of the three tasks has been well investigated,
but little literature has tried to solve them jointly.
It is hard to manually design architecture as the features from
the shared backbone now will be simultaneously affected by all the tasks.

\renewcommand{\addFig}[1]{\includegraphics[width=0.088\linewidth]{saliency_samples/#1}}
\renewcommand{\addFigs}[1]{\addFig{#1.jpg} & \addFig{#1.png} & \addFig{#1_Ours.png} &
      \addFig{#1_BAS.png} & \addFig{#1_CPD.png} & \addFig{#1_PiCA.png} & \addFig{#1_AF.png} &
      \addFig{#1_PAGE.png} & \addFig{#1_JDF.jpg} & \addFig{#1_MLMS.jpg} & \addFig{#1_DGRL.png}}
\begin{figure*}[t]
    \centering
    \footnotesize
    \renewcommand{\arraystretch}{0.8}
    \renewcommand{\tabcolsep}{0.26mm}
    \begin{tabular}{ccccccccccc} 
        \addFigs{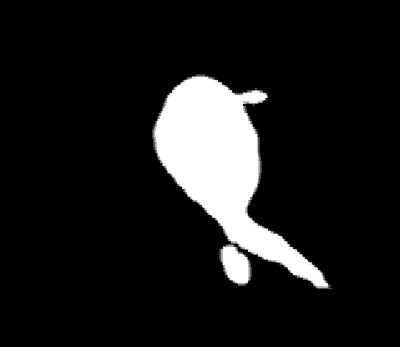} \\
        \addFigs{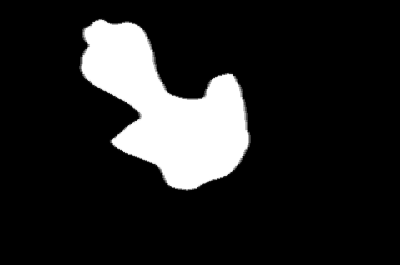} \\
        \addFigs{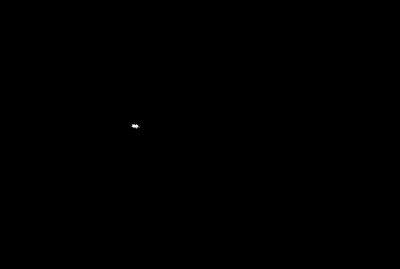} \\
        \addFigs{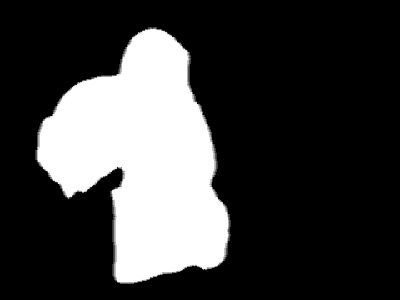} \\
        \addFigs{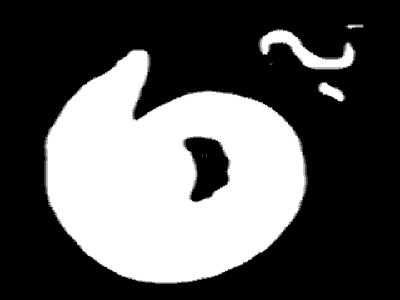} \\
        \addFigs{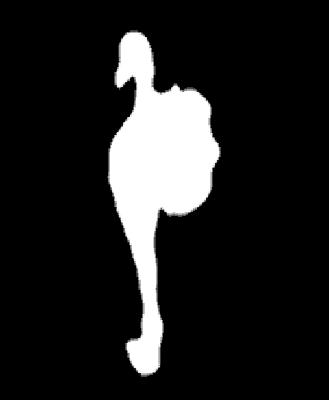}\\
        \addFigs{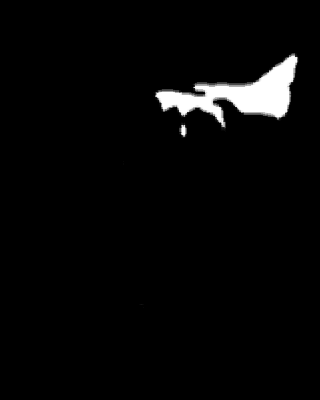}\\
        \addFigs{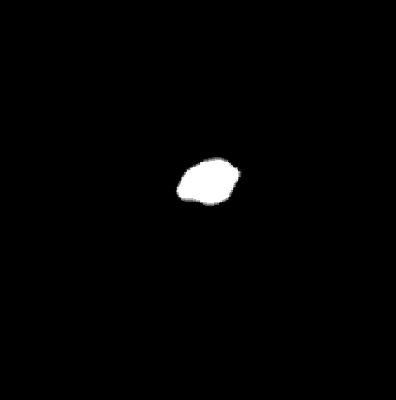}\\
        \addFigs{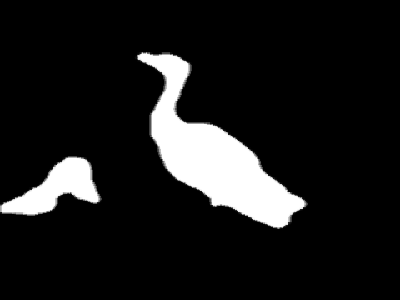}\\
        Image & GT & Ours & BASNet\cite{qin2019basnet} & CPD\cite{wu2019cascaded} & PiCANet\cite{liu2018picanet} &
        AFNet\cite{feng2019attentive} & PAGE\cite{wang2019salient} & JDFPR\cite{xu2019deepcrf} &
        MLMS\cite{wu2019mutual} & DGRL\cite{wang2018detect} \\
    \end{tabular}
    \caption{Visual comparisons of different salient object segmentation approaches.
    }\label{fig:visual_results}
\end{figure*}

\myPara{Sparse or Dense Connections}
In \tabref{tab:ablation_results}, we compare our sparse-connected
network with a dense-connected version, which means all the feature
maps in $\{S_i\}~(1\le i \le M, M=6)$ are kept rather than only
half as formulated in \eqnref{eqn:dfi}.
As shown in the 4th and 5th rows,
the dense version has worse performances on nearly all three tasks.
This may indicate that not every stage of features from the backbone
is always helpful \cite{hinton2012improving}.
For example, for edge detection, more lower-level feature maps are necessary for 
precise localization of edge pixels \cite{xie2015holistically,RcfEdgePami2019},
while for skeleton extraction, more higher-level information is essential to 
determine whether a pixel being skeleton or not \cite{zhao2018hifi,wang2019deepflux}.

\begin{table}[tp!]
  \centering
  \footnotesize
  \renewcommand{\arraystretch}{1.2}
  \renewcommand{\tabcolsep}{0.4mm}
  \caption{
    Ablation analysis of our approach on different combinations of down-sampling 
    rates of DFIMs.
    The best results in each column are highlighted in \textbf{bold}.
  }\label{tab:ablation_dfim_ds_results}
  \begin{tabular}{c|ccc|ccc|cc|c}
  \whline{1pt}
  \multirow{3}*{\shortstack{Down-\\sampling\\Rates}} & \multicolumn{6}{c|}{Saliency} & \multicolumn{2}{c|}{Edge} & Skeleton \\
  \cline{2-10}
  & \multicolumn{3}{c|}{DUT-OMRON} & \multicolumn{3}{c|}{DUTS-TE } & \multicolumn{2}{c|}{BSDS 500} & SK-LAR  \\
  \cline{2-10}
  & $F_\beta$~$\uparrow$ & MAE~$\downarrow$ & $S_m$~$\uparrow$ & $F_\beta$~$\uparrow$ & MAE~$\downarrow$ & $S_m$~$\uparrow$ & ODS~$\uparrow$ & OIS~$\uparrow$ & $F_m$~$\uparrow$ \\
  \whline{1pt}
  2,4,8 & 0.814 & 0.057 & 0.839 & 0.879 & \textbf{0.038} & 0.885 & 0.815 & 0.829 & 0.742 \\ 
  4,8,16 & 0.809 & 0.057 & 0.837 & 0.880 & \textbf{0.038} & 0.884 & 0.814 & 0.829 & \textbf{0.745}  \\
  2,4,8,16 & \textbf{0.817} & \textbf{0.056} & \textbf{0.842} & \textbf{0.884} & \textbf{0.038} & \textbf{0.887} & \textbf{0.818} & \textbf{0.834} & 0.744  \\ \hline
  \whline{1pt}
  \end{tabular}
\end{table}

\myPara{Down-Sampling Rates of DFIMs}
As listed in \tabref{tab:ablation_dfim_ds_results}, we
conduct ablation experiments on the combinations of down-sampling rates
of DFIMs.
A wider range of down-sampling rates shows a better equilibrium of overall performances,
especially on salient object segmentation and edge detection, which agrees with common
sense that richer multi-scale information usually helps.

\subsection{Task-Adaptive Attention}

\myPara{Effectiveness of TAM}
With DFIM, we can jointly train the three tasks under
a unified architecture.
However, as shown in the 5th row of \tabref{tab:ablation_results}, 
compared to separately training (the 1st row), the performance of 
skeleton extraction decreases.
As the annotations of salient object segmentation 
and edge detection tasks
pay more attention to pixels where edges exist,
which disagree with the goal of skeleton extraction task,
the optimization
of skeleton extraction task could be influenced and misguided
towards adverse directions.
With TAM, the network is able to allocate
the information of all tasks from a global view by adjusting
the gradients of each task towards the shared backbone adaptively.
As can be seen from the 7th row compared to the 5th row in \tabref{tab:ablation_results},
better overall performances are reached.
The performances of salient object segmentation and edge detection are slightly
better while the skeleton extraction task outperforms with 0.7\%.

\begin{table*}[tp!]
  \centering
  \footnotesize
  \renewcommand{\arraystretch}{1.15}
  \renewcommand{\tabcolsep}{0.75mm}
  \caption{Quantitative salient object segmentation results six widely used datasets.
  The best result in each column are highlighted in \textbf{bold}.
  As can be seen, our approach achieves the best results on nearly all datasets in terms of F-measure, MAE and S-measure.
  }\label{tab:results}
  \begin{tabular}{l|ccc|ccc|ccc|ccc|ccc|ccc}
  \whline{1pt}
   & \multicolumn{3}{c}{ECSSD \cite{yan2013hierarchical}} & \multicolumn{3}{c}{PASCAL-S \cite{li2014secrets}} & \multicolumn{3}{c}{DUT-OMRON \cite{yang2013saliency}} & \multicolumn{3}{c}{HKU-IS \cite{li2015visual}} & \multicolumn{3}{c}{SOD \cite{movahedi2010design}} & \multicolumn{3}{c}{DUTS-TE \cite{wang2017learning}} \\
   \cline{2-4} \cline{5-7} \cline{8-10} \cline{11-13} \cline{14-16} \cline{17-19}
   Model & $F_\beta$~$\uparrow$ & MAE~$\downarrow$ & $S_m$~$\uparrow$ & $F_\beta$~$\uparrow$ & MAE~$\downarrow$ & $S_m$~$\uparrow$ & $F_\beta$~$\uparrow$ & MAE~$\downarrow$ & $S_m$~$\uparrow$ & $F_\beta$~$\uparrow$ & MAE~$\downarrow$ & $S_m$~$\uparrow$ & $F_\beta$~$\uparrow$ & MAE~$\downarrow$ & $S_m$~$\uparrow$ & $F_\beta$~$\uparrow$ & MAE~$\downarrow$ & $S_m$~$\uparrow$ \\
  \whline{1pt} 
  $\text{DCL}_\text{16}$~\cite{li2016deep} & 0.896 & 0.080 & 0.869 & 0.805 & 0.115 & 0.800 & 0.733 & 0.094 & 0.762 & 0.893 & 0.063 & 0.871 & 0.831 & 0.131 & 0.763 & 0.786 & 0.081 & 0.803 \\
  $\text{RFCN}_\text{16}$~\cite{wangsaliency} & 0.898 & 0.097 & 0.856 & 0.827 & 0.118 & 0.808 & 0.747 & 0.094 & 0.774 & 0.895 & 0.079 & 0.860 & 0.805 & 0.161 & 0.722 & 0.786 & 0.090 & 0.793 \\
  $\text{MSR}_\text{17}$~\cite{li2017instance} & 0.903 & 0.059 & 0.887 & 0.839 & 0.083 & 0.835 & 0.790 & 0.073 & 0.805 & 0.907 & 0.043 & 0.896 & 0.841 & 0.111 & 0.782 & 0.824 & 0.062 & 0.834 \\
  $\text{DSS}_\text{17}$~\cite{hou2016deeply} & 0.906 & 0.064 & 0.880 & 0.821 & 0.101 & 0.804 & 0.760 & 0.074 & 0.789 & 0.900 & 0.050 & 0.881 & 0.834 & 0.125 & 0.764 & 0.813 & 0.065 & 0.826 \\
  $\text{NLDF}_\text{17}$~\cite{luo2017non} & 0.903 & 0.065 & 0.870 & 0.822 & 0.098 & 0.805 & 0.753 & 0.079 & 0.770 & 0.902 & 0.048 & 0.878 & 0.837 & 0.123 & 0.759 & 0.816 & 0.065 & 0.816 \\
  $\text{Amulet}_\text{17}$~\cite{zhang2017amulet} & 0.911 & 0.062 & 0.876 & 0.826 & 0.092 & 0.816 & 0.737 & 0.083 & 0.784 & 0.889 & 0.052 & 0.866 & 0.799 & 0.146 & 0.729 & 0.773 & 0.075 & 0.800 \\
  $\text{PAGR}_\text{18}$~\cite{zhang2018progressive} & 0.924 & 0.064 & 0.883 & 0.847 & 0.089 & 0.822 & 0.771 & 0.071 & 0.775 & 0.919 & 0.047 & 0.889 & - & - & - & 0.854 & 0.055 & 0.839 \\
  $\text{DGRL}_\text{18}$~\cite{wang2018detect} & 0.921 & 0.043 & 0.899 & 0.844 & 0.072 & 0.836 & 0.774 & 0.062 & 0.806 & 0.910 & 0.036 & 0.895 & 0.843 & 0.103 & 0.774 & 0.828 & 0.049 & 0.842 \\
  $\text{MLMS}_\text{19}$~\cite{wu2019mutual} & 0.924 & 0.048 & 0.905 & 0.853 & 0.074 & 0.844 & 0.793 & 0.063 & 0.809 & 0.922 & 0.039 & 0.907 & 0.857 & 0.106 & 0.790 & 0.854 & 0.048 & 0.862 \\ 
  $\text{JDFPR}_\text{19}$~\cite{xu2019deepcrf} & 0.925 & 0.052 & 0.902 & 0.854 & 0.082 & 0.841 & 0.802 & 0.057 & 0.821 & - & - & - & 0.836 & 0.121 & 0.767 & 0.833 & 0.058 & 0.836 \\ 
  $\text{PAGE}_\text{19}$~\cite{wang2019salient} & 0.928 & 0.046 & 0.906 & 0.848 & 0.076 & 0.842 & 0.791 & 0.062 & 0.825 & 0.920 & 0.036 & 0.904 & 0.837 & 0.110 & 0.775 & 0.838 & 0.051 & 0.855 \\ 
  $\text{CapSal}_\text{19}$~\cite{zhang2019capsal} & - & - & - & 0.862 & 0.073 & 0.837 & - & - & - & 0.889 & 0.058 & 0.851 & - & - & - & 0.844 & 0.060 & 0.818 \\ 
  $\text{CPD}_\text{19}$~\cite{wu2019cascaded} & 0.936 & 0.040 & 0.913 & 0.859 & 0.071 & 0.848 & 0.796 & 0.056 & 0.825 & 0.925 & 0.034 & 0.907 & 0.857 & 0.110 & 0.771 & 0.865 & 0.043 & 0.869 \\ 
  $\text{PiCANet}_\text{18}$~\cite{liu2018picanet} & 0.932 & 0.048 & 0.912 & 0.864 & 0.075 & 0.854 & 0.820 & 0.064 & 0.830 & 0.920 & 0.044 & 0.904 & 0.861 & 0.103 & 0.792 & 0.863 & 0.050 & 0.868 \\ 
  $\text{AFNet}_\text{19}$~\cite{feng2019attentive} & 0.932 & 0.045 & 0.907 & 0.861 & 0.070 & 0.849 & 0.820 & 0.057 & 0.825 & 0.926 & 0.036 & 0.906 & - & - & - & 0.867 & 0.045 & 0.867 \\ 
  $\text{BASNet}_\text{19}$~\cite{qin2019basnet} & 0.939 & 0.040 & 0.911 & 0.857 & 0.076 & 0.838 & 0.811 & 0.057 & 0.836 & 0.930 & 0.033 & 0.908 & 0.849 & 0.112 & 0.772 & 0.860 & 0.047 & 0.866 \\ \hline
  \textbf{DFI (Ours)} & \textbf{0.945} & \textbf{0.038} & \textbf{0.921} & \textbf{0.880} & \textbf{0.065} & \textbf{0.865} & \textbf{0.829} & \textbf{0.055} & \textbf{0.839} & \textbf{0.934} & \textbf{0.031} & \textbf{0.919} & \textbf{0.878} & \textbf{0.100} & \textbf{0.802} & \textbf{0.888} & \textbf{0.038} & \textbf{0.887} \\
  \whline{1pt}
  \end{tabular}
\end{table*}

\myPara{Necessity of Information Exchange}
To investigate whether the promotion brought 
by TAM is due to the introduction of additional parameters,
we also conduct experiment leaving the parameters of 
different branches in TAM unshared (the 6th row) so that different task branches
are independent of each other after selecting features from the shared backbone.
By not sharing the parameters in TAM, extra 1.66M parameters are further lead in.
However, 
as can be seen from the 6th row in \tabref{tab:ablation_results}, 
even with more parameters introduced,
the overall performance of the unshared version is obviously 
worse than the shared version of TAM (the 7th row).
Though the skeleton task performs slightly better, the other two tasks decline greatly.
These phenomenons show that enforcing the interchange of information across tasks 
after the separation of branches of each task
is helpful
to the overall convergence of all tasks, 
while simple attention mechanism works not well.
This can also be observed from the first two rows of \tabref{tab:ablation_results},
that appending TAM when each task is trained separately brings no help even
downgrade to most of the three tasks.

\renewcommand{\addFig}[1]{\includegraphics[width=0.31\linewidth]{prs/#1.pdf}}
\begin{figure*}[t]
  \centering
  \footnotesize
  \setlength\tabcolsep{1.4mm}
  \includegraphics[width=.95\linewidth]{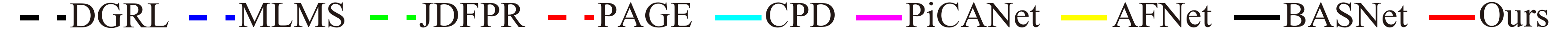}
  \renewcommand\arraystretch{1.2}
  \begin{tabular}{cccc}
    \addFig{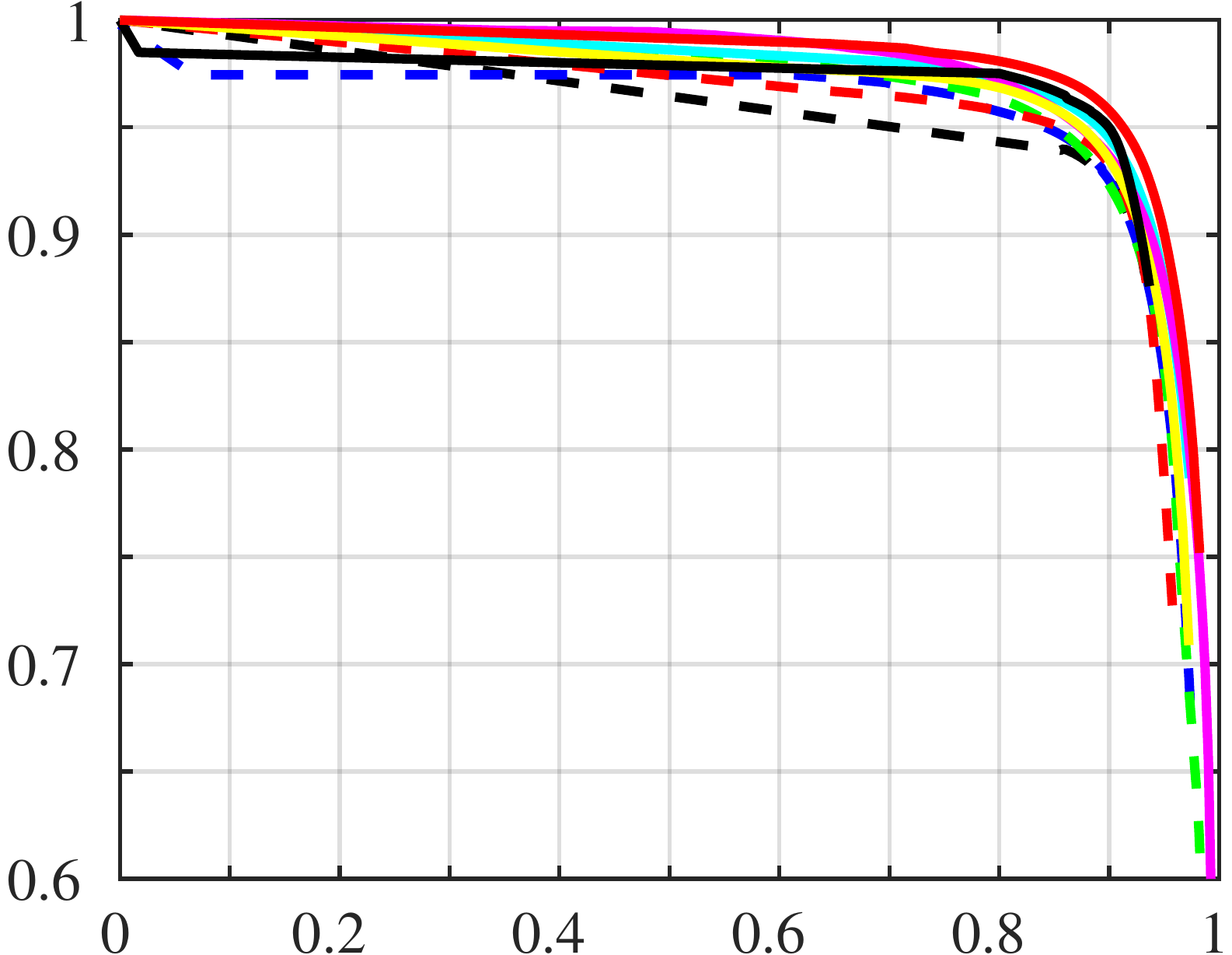} & \addFig{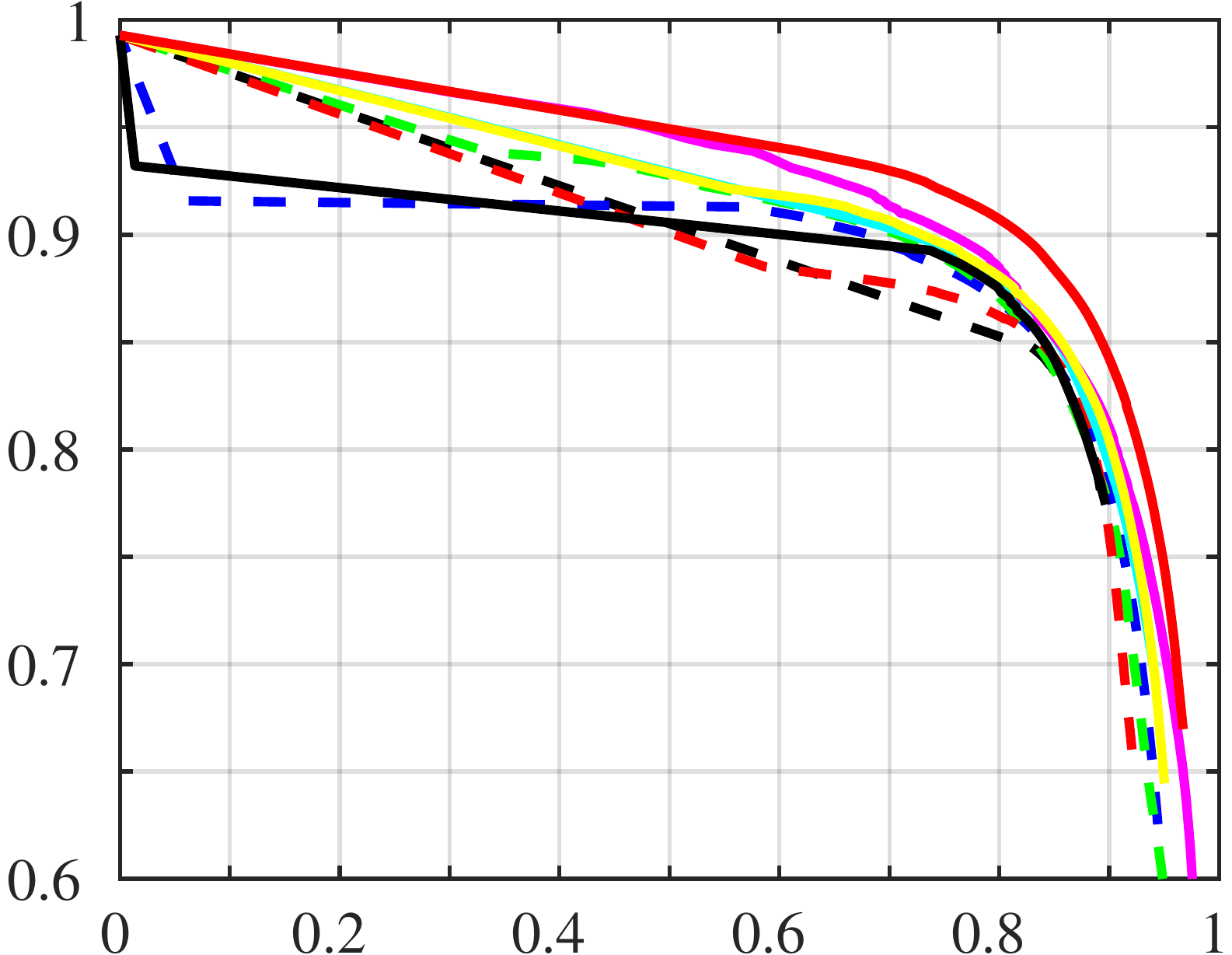} & \addFig{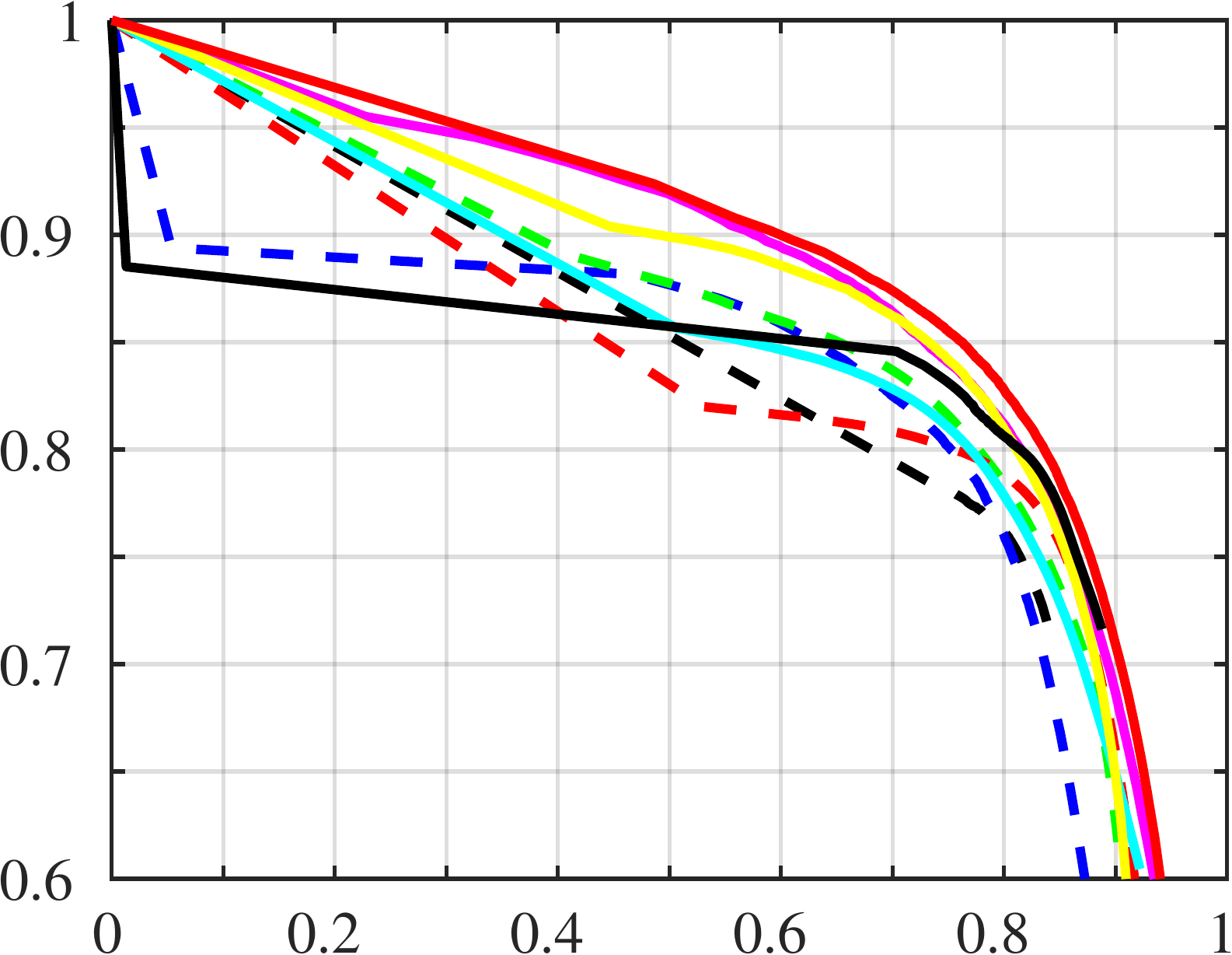} \\
    ~~~~(a) ECSSD~\cite{yan2013hierarchical}  & ~~~~(b) PASCAL-S~\cite{li2014secrets} & ~~~~(c) DUT-OMRON~\cite{yang2013saliency} \\
    \addFig{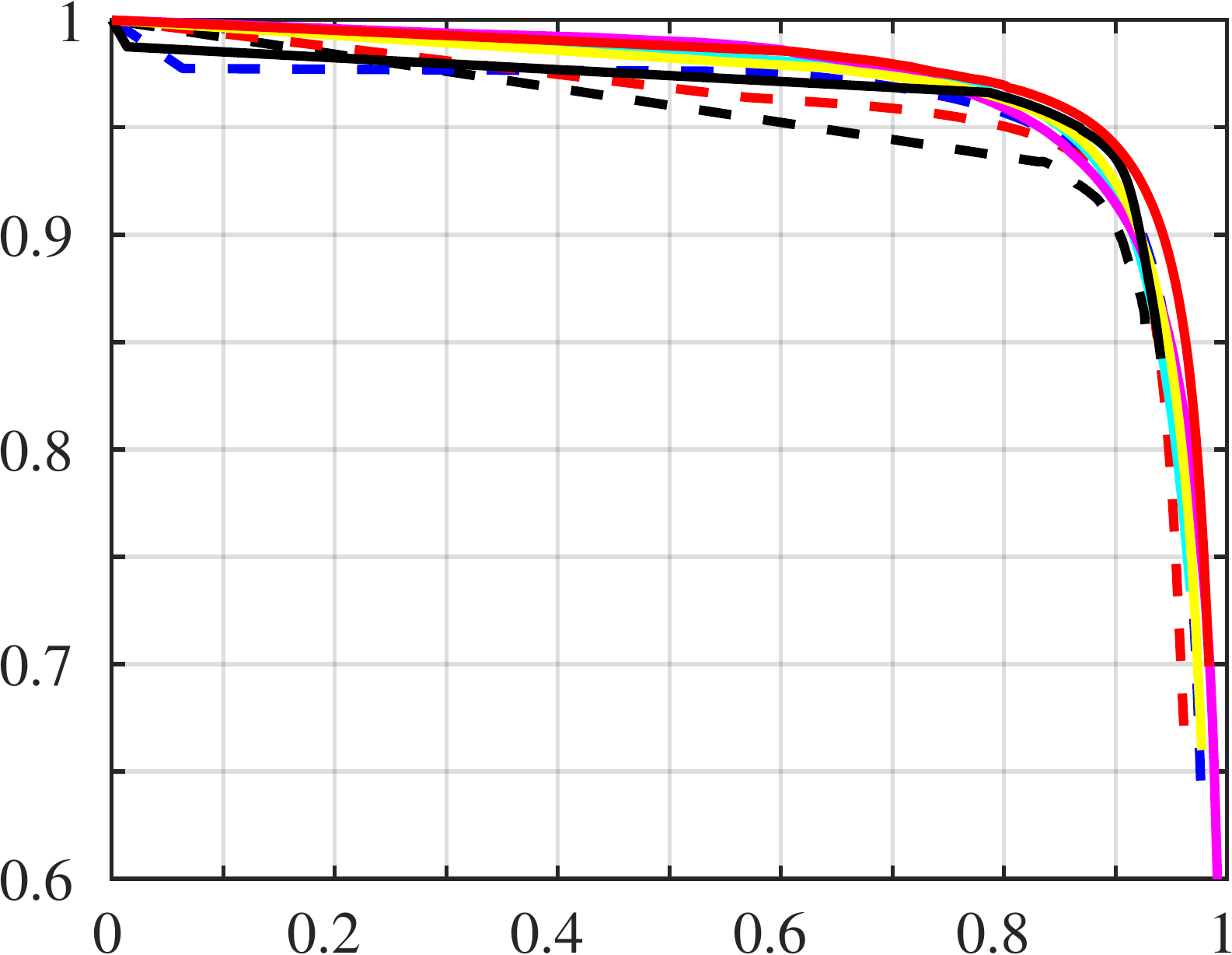} & \addFig{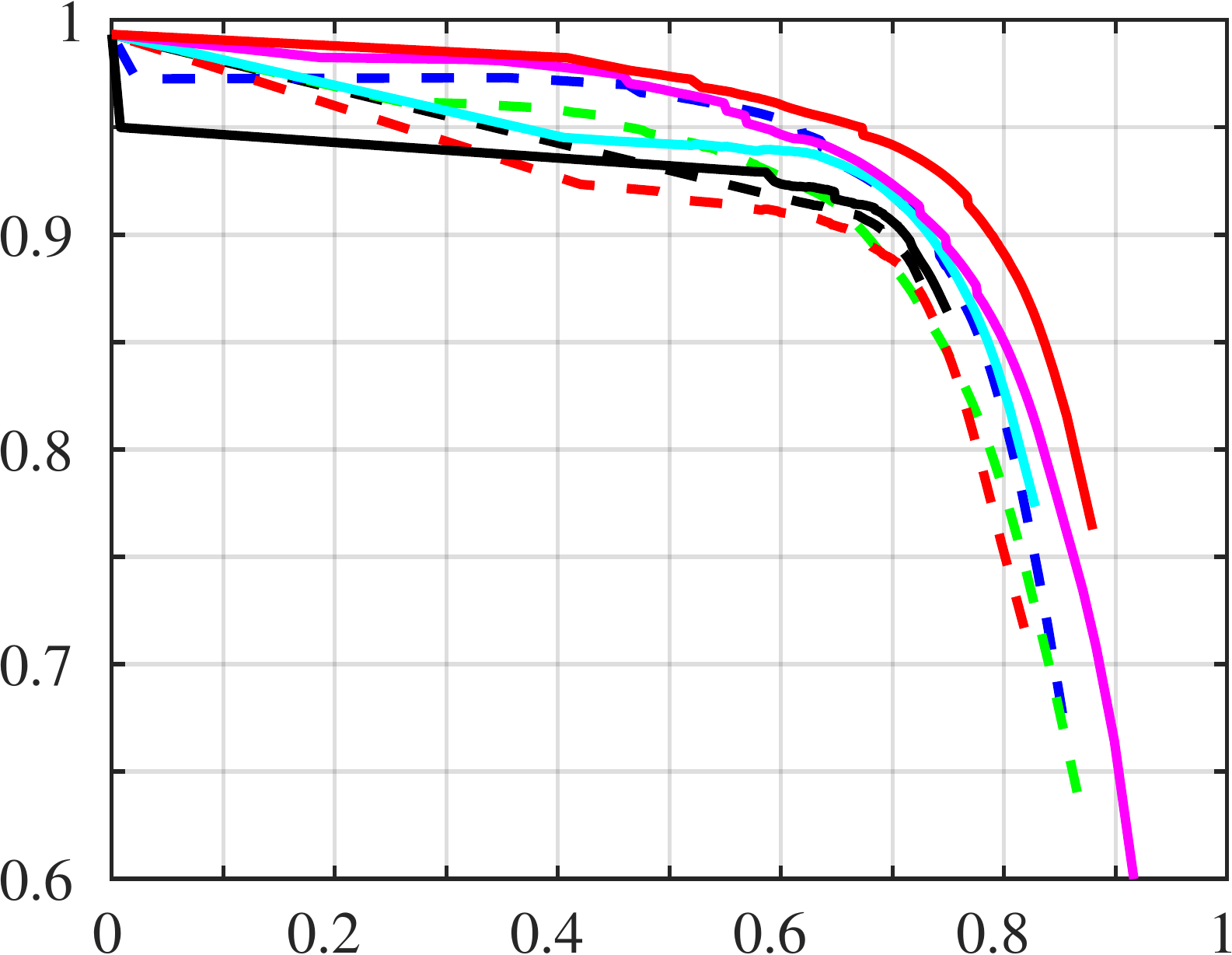} & \addFig{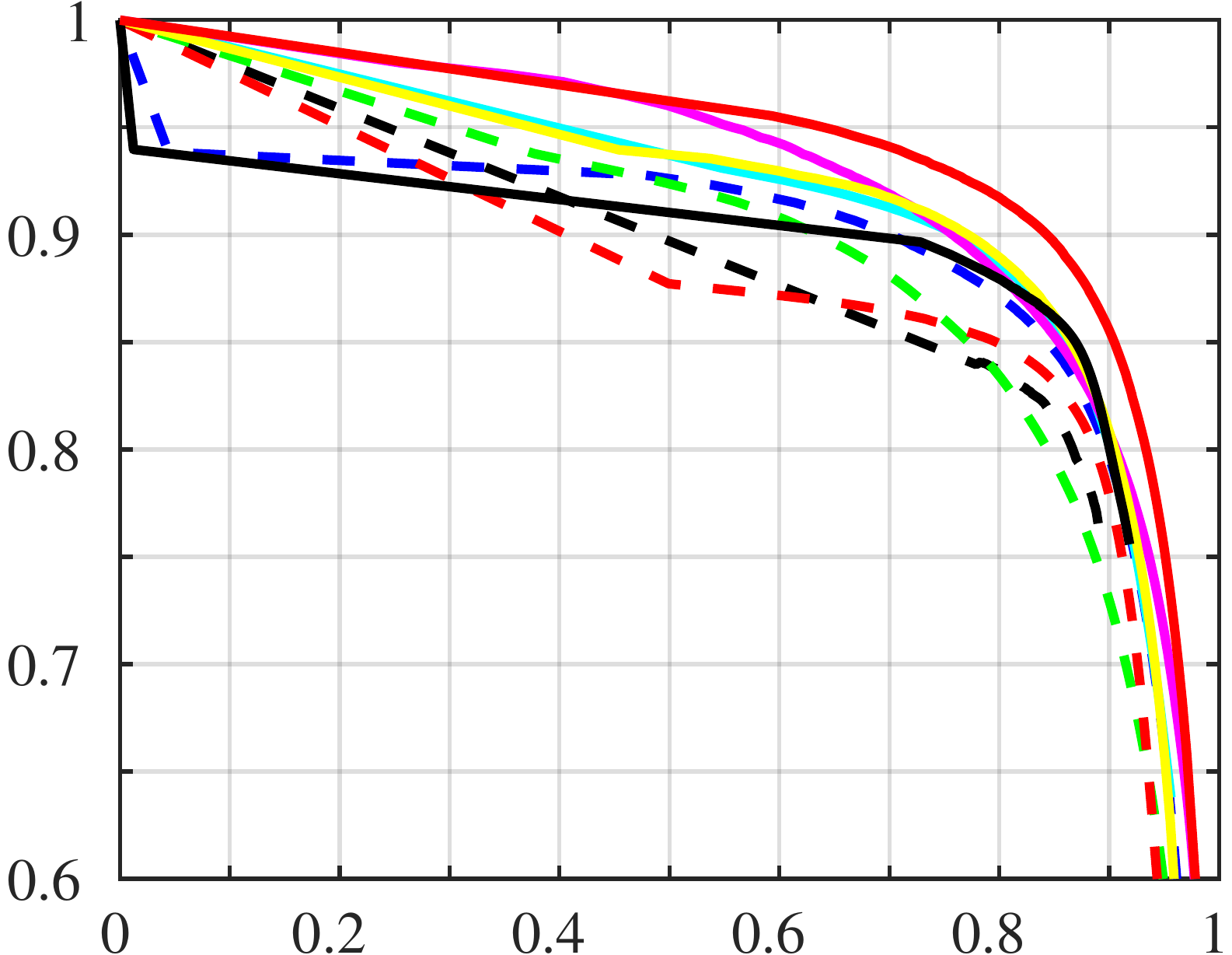} \\
    ~~~~(d) HKU-IS~\cite{li2015visual} & ~~~~(e) SOD~\cite{movahedi2010design} & ~~~~(f) DUTS-TE~\cite{wang2017learning} \\
  \end{tabular}
  \caption{Precision (vertical axis) recall (horizontal axis) curves on
    six popular salient object segmentation datasets.}
  \label{fig:salient_prs}
\end{figure*}

\section{Comparisons to the State-of-the-Arts} \label{sec:sota}
In this section, we compare the proposed method (denoted as DFI for convenience) with
\sArt methods on salient object segmentation, 
edge detection, and skeleton extraction.
As very little literature has solved the three tasks jointly before,
\eg UberNet \cite{kokkinos2017ubernet} (CVPR'17) and 
MLMS \cite{wu2019mutual} (CVPR'19), which solved salient object segmentation jointly
with edge detection, 
we mainly compare with the \sArt single-purpose methods of the three tasks
for better illustration.
For fair comparisons, for each task, the predicted maps 
(\eg saliency maps, edge maps, skeleton maps) of other methods
are generated by the original code released by the authors or
directly provided by them.
All the results are obtained directly from single-model test
without relying on any other pre- or post-processing tools except for the NMS 
process before the evaluation of 
edge and skeleton maps \cite{xie2015holistically,RcfEdgePami2019,zhao2018hifi}.
And for each task, all the predicted maps are evaluated with the same evaluation code.

\subsection{Salient Object Segmentation}
We exhaustively compare DFI with 16 existing state-of-the-art salient object segmentation
methods including DCL~\cite{li2016deep}, RFCN~\cite{wangsaliency}, MSR~\cite{li2017instance}, DSS~\cite{hou2016deeply}, NLDF~\cite{luo2017non},
Amulet~\cite{zhang2017amulet}, PAGR~\cite{zhang2018progressive}, DGRL~\cite{wang2018detect},
MLMS \cite{wu2019mutual}, JDFPR~\cite{xu2019deepcrf},
PAGE \cite{wang2019salient}, CapSal~\cite{zhang2019capsal}, CPD~\cite{wu2019cascaded}, PiCANet~\cite{liu2018picanet},
AFNet \cite{feng2019attentive}, and BASNet~\cite{qin2019basnet}.
%

\begin{figure}[tp]
  \centering
  \footnotesize
  \renewcommand{\tabcolsep}{2mm}
  \begin{tabular}{c}
      \includegraphics[width=0.75\linewidth]{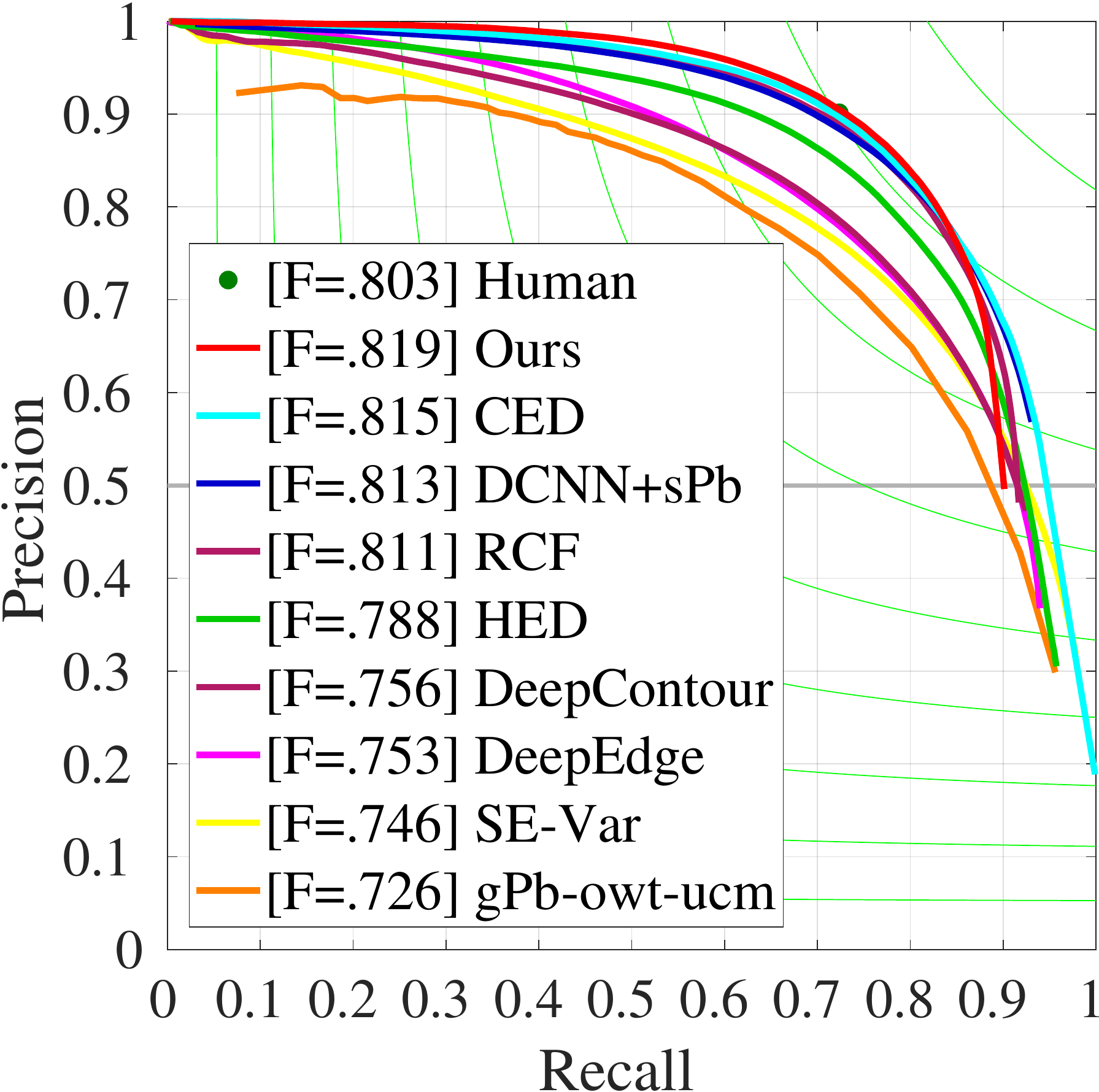}
  \end{tabular}
  \caption{The precision-recall curves on BSDS 500 dataset \cite{arbelaez2011contour}.}
  \label{fig:edge_pr}
\end{figure}

\myPara{F-measure, MAE and S-measure Scores}
Here, we compare DFI with the aforementioned approaches in terms of F-measure, MAE, and
S-measure (See \tabref{tab:results}).
%
%
As can be seen, compared to the second-best methods on each dataset,
DFI outperforms all of them over six datasets
with average promotions of 1.2\% and 1.0\%
in terms of F-measure and S-measure, respectively.
Especially on the challenging DUTS-TE dataset, promotions of 2.1\%
and 1.8\% in terms of F-measure and S-measure can be observed.
Similar patterns can also be observed using the MAE score.
Also, when compared to MLMS \cite{wu2019mutual}, which 
learns salient object segmentation and 
edge detection jointly, DFI has even larger improvements on both tasks,
as shown in the 7th and 9th rows of \tabref{tab:ablation_results}.
Without TAM, DFI still outperforms MLMS \cite{wu2019mutual} 
by a large margin (the 3rd and 9th rows).
This phenomenon demonstrated the effectiveness of the proposed DFIM and TAM,

\renewcommand{\addFig}[1]{\includegraphics[width=0.163\linewidth]{edge_samples/#1}}
\renewcommand{\addFigs}[1]{\addFig{#1.jpg} & \addFig{#1_gt.png} & \addFig{#1_dfi.png} & \addFig{#1_ced.png} &
    \addFig{#1_rcf.png} & \addFig{#1_hed.png}}
\begin{figure*}[tp]
     \centering
     \footnotesize
     \renewcommand{\tabcolsep}{0.3mm}
     \begin{tabular*}{1.0\textwidth}{cccccc}
          \addFigs{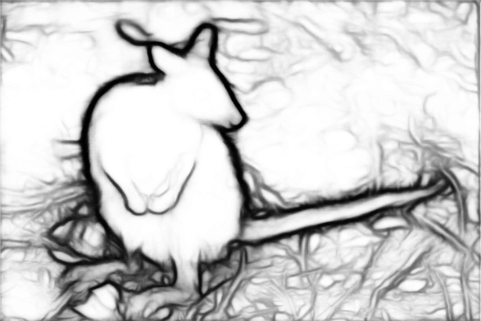} \\
          \addFigs{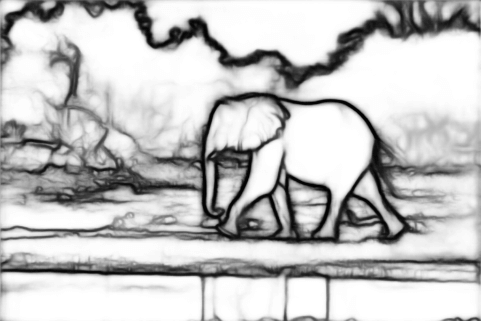} \\
          \addFigs{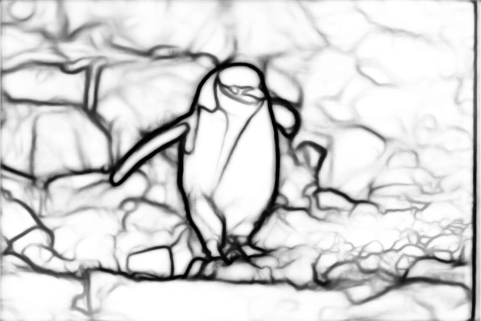} \\
          \addFigs{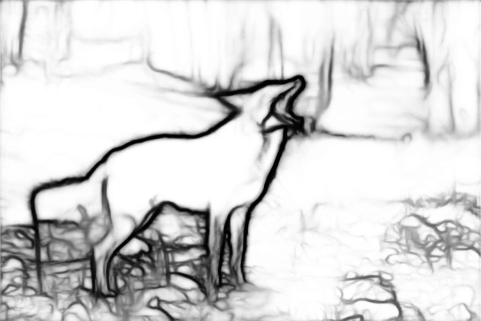} \\
          (a) Image & (b) GT & (c) Ours & (d) CED \cite{wang2017deep} 
          & (e) RCF \cite{RcfEdgePami2019}& (f) HED \cite{xie2015holistically}
     \end{tabular*}
     \caption{Visual comparisons with several recent state-of-the-art edge detectors. 
     As can be seen, DFI is able to
     not only generate cleaner background but also more confident on the 
     object boundaries compared to the other methods.}
     \label{fig:edge_vis}
 \end{figure*}

\myPara{PR Curves}
Other than numerical results, we also show
the PR curves on the six datasets 
as shown in \figref{fig:salient_prs}.
As can be seen, the PR curves of DFI (red solid ones) 
are comparable to other previous 
approaches and even better on some datasets.
Especially on the PASCAL-S and DUTS-TE datasets, 
DFI outstands compared to all other previous approaches. 
As the recall score approaches 1, our precision
score is much higher than other methods,
which reveals that the false positives in our saliency map are low.

\begin{table}[t]
  \centering
  \small
  \renewcommand{\arraystretch}{1.1}
  \renewcommand{\tabcolsep}{3mm}
  \caption{Quantitative comparisons of DFI with existing edge detection methods.
  The best results in each column are highlighted in \textbf{bold}.}\label{tab:edge_comps}
  \begin{tabular}{l||cc}
    \whline{1pt}
    & \multicolumn{2}{c}{BSDS 500 \cite{arbelaez2011contour}}  \\ \cline{2-3}
    Method & ODS~$\uparrow$ & OIS~$\uparrow$ \\ \whline{1pt}
      $\text{gPb-owt-ucm}_\text{11}$ \cite{arbelaez2011contour}  & 0.726 & 0.757 \\ 
      $\text{SE-Var}_\text{15}$ \cite{dollar2015fast}  & 0.746 & 0.767 \\ 
      $\text{MCG}_\text{17}$ \cite{pont2017multiscale}  & 0.747 & 0.779 \\ \hline
      $\text{DeepEdge}_\text{15}$ \cite{bertasius2015deepedge}  & 0.753 & 0.772 \\  
      $\text{DeepContour}_\text{15}$ \cite{shen2015deepcontour}  & 0.756 & 0.773 \\ 
      $\text{HED}_\text{15}$ \cite{xie2015holistically}  & 0.788 & 0.808 \\ 
      $\text{CEDN}_\text{16}$ \cite{yang2016object}  & 0.788 & 0.804 \\ 
      $\text{RDS}_\text{16}$ \cite{liu2016learning} & 0.792 & 0.810 \\  
      $\text{COB}_\text{17}$ \cite{maninis2017convolutional} & 0.793 & 0.820 \\  
      $\text{RCF}_\text{17}$ \cite{RcfEdgePami2019}  & 0.811 & 0.830 \\ 
      $\text{DCNN+sPb}_\text{15}$ \cite{kokkinos2015pushing}  & 0.813 & 0.831 \\ 
      $\text{CED}_\text{17}$ \cite{wang2017deep} & 0.815 & 0.833 \\ 
      $\text{LPCB}_\text{18}$ \cite{deng2018learning} & 0.815 & 0.834 \\ \hline
      \textbf{DFI (Ours)} & \textbf{0.819} & \textbf{0.836} \\
      \whline{1pt}
      \end{tabular}
\end{table}

\begin{figure*}[tp]
  \centering
  \footnotesize
  \renewcommand{\arraystretch}{1.4}
  \renewcommand{\tabcolsep}{7mm}
  \begin{tabular}{cc}
      \includegraphics[width=0.35\linewidth]{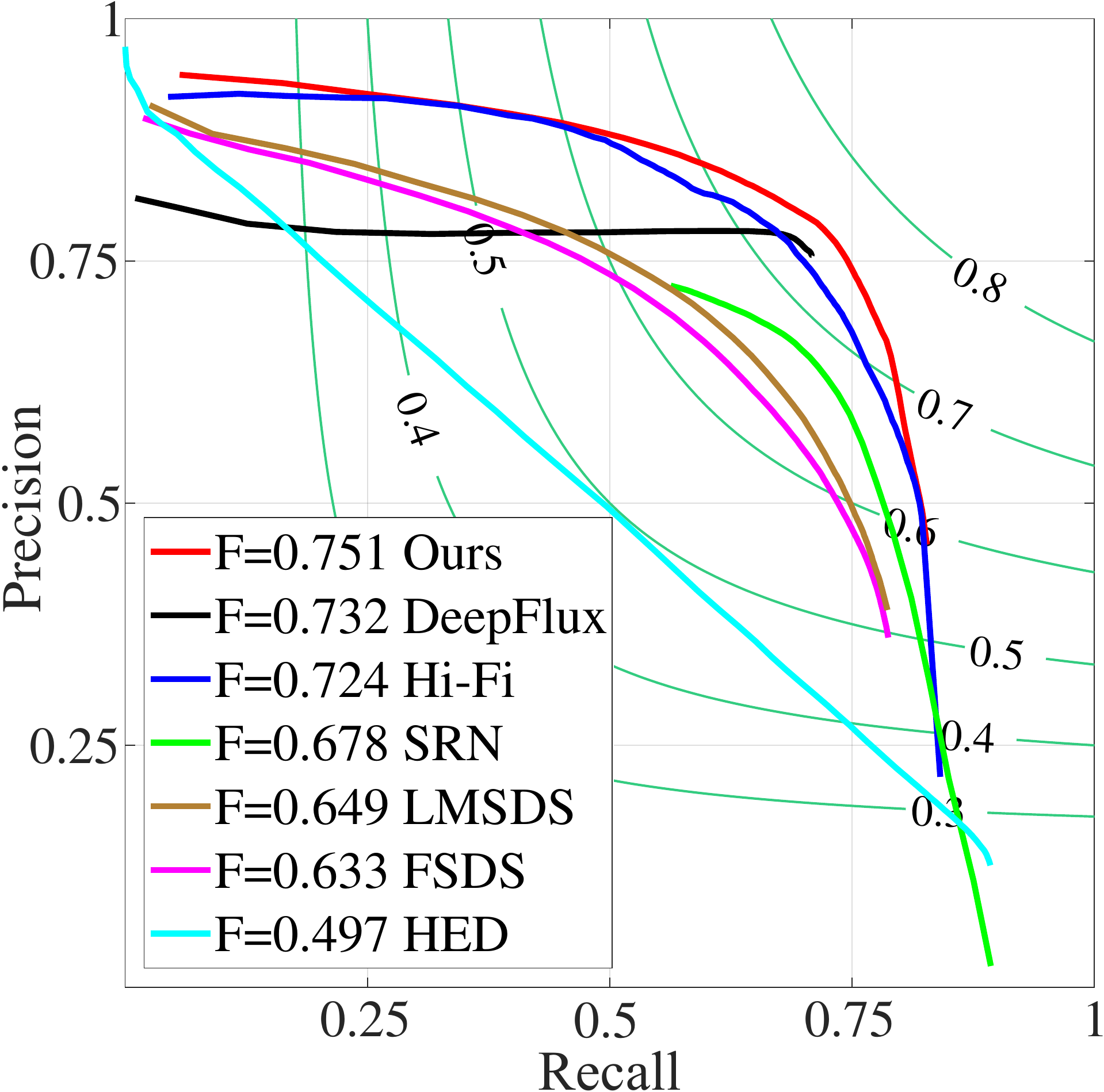}~~~~ &
      \includegraphics[width=0.35\linewidth]{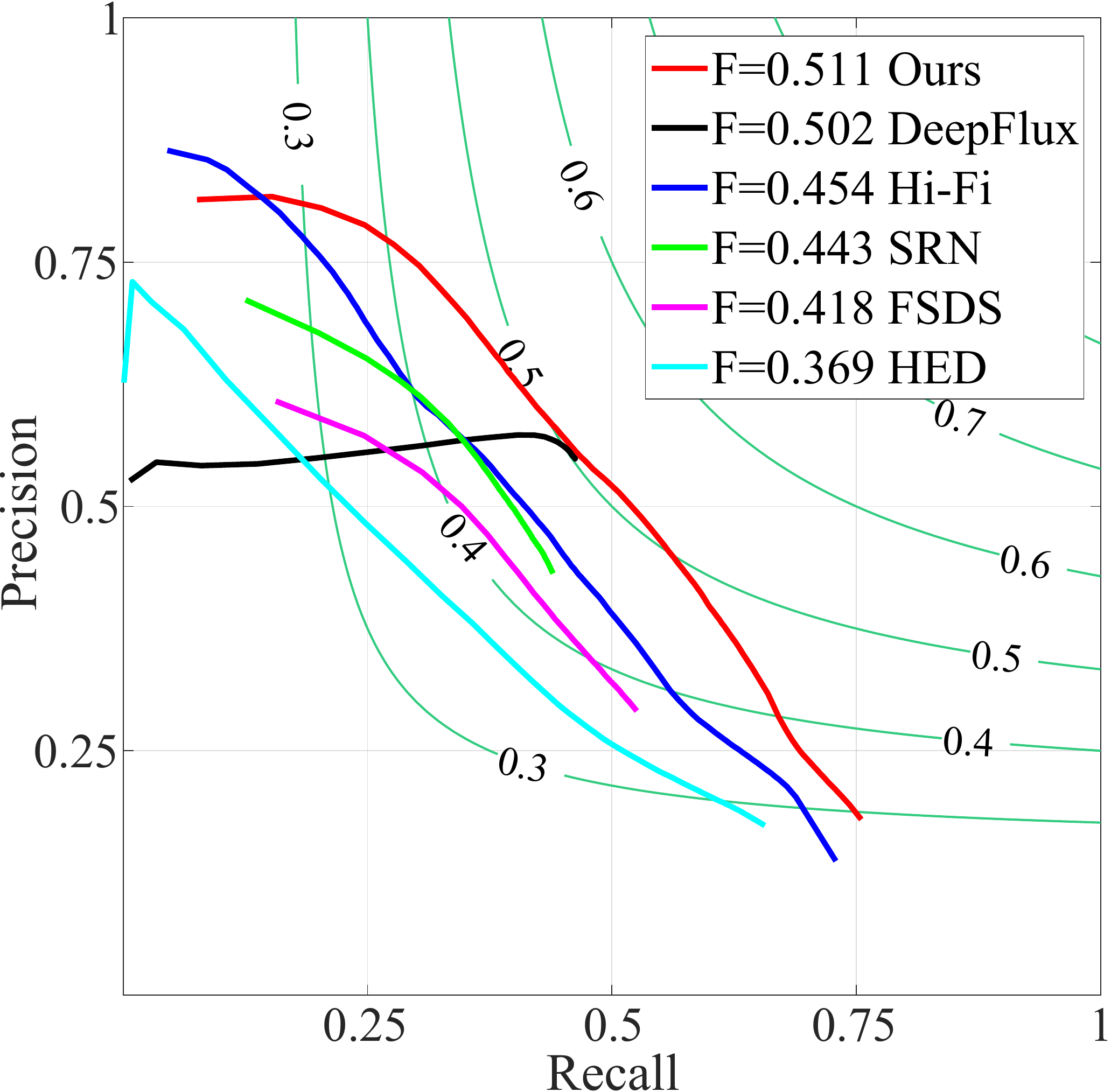} \\
      ~~~~(a) SK-LARGE~\cite{shen2017deepskeleton} & ~~~~(b) SYM-PASCAL~\cite{ke2017srn} \\
  \end{tabular}
  \caption{The precision-recall curves of some selected skeleton extraction methods
  on SK-LARGE dataset \cite{shen2017deepskeleton}
  and SYM-PASCAL dataset \cite{ke2017srn}.}
  \label{fig:skel_pr}
\end{figure*}

\begin{table}[t]
  \centering
  \small
  \renewcommand{\arraystretch}{1.1}
  \renewcommand{\tabcolsep}{1.5mm}
  \caption{Quantitative comparisons of DFI with existing skeleton extraction methods.
  The best results in each column are highlighted in \textbf{bold}.}\label{tab:skel_comps}
  \begin{tabular}{l||cc}
    \whline{1pt}
    & SK-LARGE \cite{shen2017deepskeleton} & SYM-PASCAL \cite{ke2017srn} \\ 
    \cline{2-2} \cline{3-3}
    Method & $F_m$~$\uparrow$ & $F_m$~$\uparrow$ \\ \whline{1pt}
      $\text{MIL}_\text{12}$ \cite{tsogkas2012learning} & 0.353 & 0.174 \\ 
      $\text{HED}_\text{15}$ \cite{xie2015holistically} & 0.497 & 0.369 \\ 
      $\text{RCF}_\text{17}$ \cite{RcfEdgePami2019} & 0.626 & 0.392 \\ 
      $\text{FSDS}_\text{16}$ \cite{shen2016object} & 0.633 & 0.418 \\ 
      $\text{LMSDS}_\text{17}$ \cite{shen2017deepskeleton} & 0.649 & - \\ 
      $\text{SRN}_\text{17}$ \cite{ke2017srn} & 0.678 & 0.443 \\ 
      $\text{LSN}_\text{18}$ \cite{liu2018linear} & 0.668 & 0.425 \\ 
      $\text{Hi-Fi}_\text{18}$ \cite{zhao2018hifi} & 0.724 & 0.454 \\ 
      $\text{DeepFlux}_\text{19}$ \cite{wang2019deepflux} & 0.732 & 0.502 \\ \hline
      \textbf{DFI (Ours)} & \textbf{0.751} & \textbf{0.511} \\ 
      \whline{1pt}
      \end{tabular}
\end{table}

\myPara{Visual Comparisons}
In \figref{fig:visual_results}, we show the visual comparisons with several previous state-of-the-art
approaches. 
In the top row, the salient object is partially occluded.
And DFI is able to segment the entire object without 
mixing in the unrelated areas.
As shown in the 2nd row,
DFI is also able to segment out the salient object with more precise boundaries and details.
A similar phenomenon happens when processing images where salient objects are tiny and irregular
or the contrast between foreground and
background is low. 
%
%
For example, the bottom two rows of \figref{fig:visual_results}. 
These results demonstrate that DFI benefits from better distinguishing the 
edge pixels and segmenting out the whole objects,
which might be the advantage of joint training with the edge detection and skeleton extraction tasks.

\renewcommand{\addFig}[1]{\includegraphics[width=0.19\linewidth]{skeleton_samples/#1}}
\renewcommand{\addFigs}[1]{\addFig{#1.png} & \addFig{#1_DFI_mk.png} & \addFig{#1_Hi-Fi_mk.png} &
    \addFig{#1_SRN_mk.png} & \addFig{#1_FSDS_mk.png}}
\begin{figure}[!t]
  \centering
  \scriptsize
  \renewcommand{\tabcolsep}{0.3mm}
  \begin{tabular}{ccccc}
     \addFigs{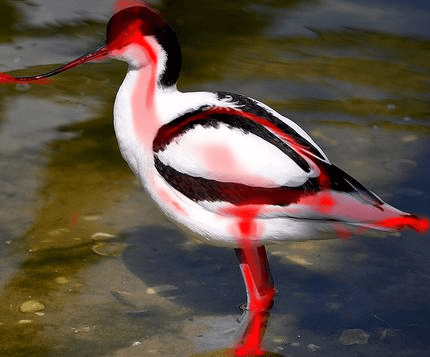} \\
     \addFigs{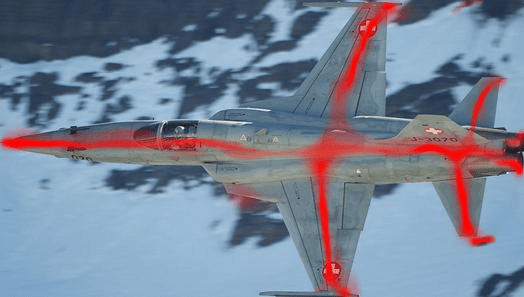} \\
     \addFigs{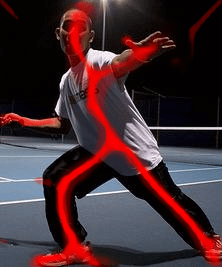} \\
     \addFigs{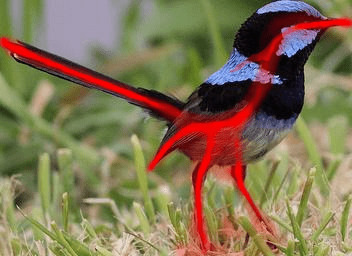} \\
     \addFigs{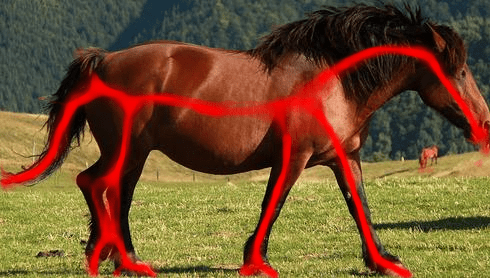} \\
     \addFigs{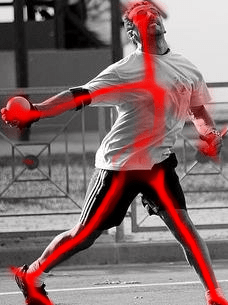} \\
     \addFigs{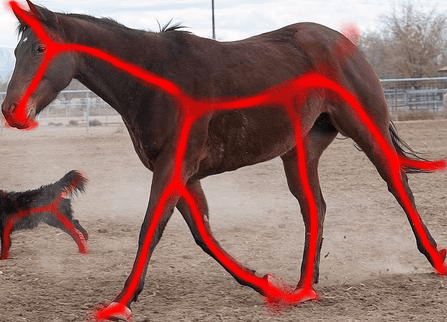} \\
     \addFigs{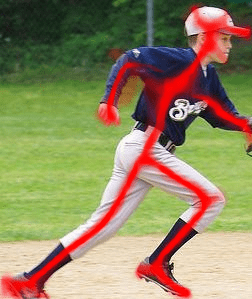} \\
     \addFigs{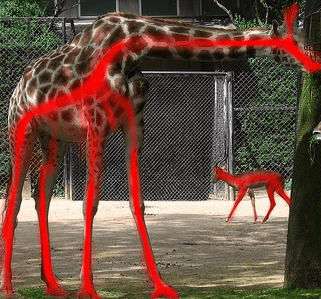} \\
     (a) GT & (b) Ours & (c) Hi-Fi \cite{zhao2018hifi} 
     & (d) SRN \cite{ke2017srn} & (e) FSDS \cite{shen2016object} 
  \end{tabular}
  \caption{Visual comparisons with three recently representative skeleton extraction approaches.
  It can be easily found that our results are much thinner and stronger than the other three methods. Also,
  the skeletons produced by our results are continuous, which is essential for their applications.}
  \label{fig:sk_vis}
\end{figure}

\subsection{Edge Detection}
We compare DFI with results from 13 existing \sArt edge detection methods, 
including gPb-owt-ucm \cite{arbelaez2011contour}, SE-Var \cite{dollar2015fast}, MCG \cite{pont2017multiscale}, 
DeepEdge \cite{bertasius2015deepedge}, DeepContour \cite{shen2015deepcontour}, HED \cite{xie2015holistically}, 
CEDN \cite{yang2016object}, RDS \cite{liu2016learning}, COB \cite{maninis2017convolutional}, 
RCF \cite{RcfEdgePami2019}, DCNN+sPb \cite{kokkinos2015pushing}, CED \cite{wang2017deep} and LPCB \cite{deng2018learning}, most of which are CNN-based methods.

\myPara{Quantitative Analysis} In \tabref{tab:edge_comps}, we show the quantitative results.
%
DFI achieves ODS of 0.819 and OIS of 0.836, which are
even better than the previous works that are well-designed for
edge detection.
Thanks to DFIM and TAM,
the information from the
other tasks not only does not influence but helps the 
performance of edge detection, as shown in the 1st and 7th rows of
the `Edge' column of \tabref{tab:ablation_results}.

\myPara{PR Curves}
The precision-recall curves of our method and some selected methods on the 
BSDS 500 dataset \cite{arbelaez2011contour} can be found in \figref{fig:edge_pr}.
One can observe that the PR curve produced by our approach is already better than
human in some certain cases and is comparable to previous methods especially in precision.

\myPara{Visual Analysis} In \figref{fig:edge_vis}, we show some visual comparisons 
between DFI and some leading representative methods \cite{wang2017deep,liu2016learning,xie2015holistically}.
As can be observed, DFI performs better in detecting the boundaries compared to the others.
In the last row of \figref{fig:edge_vis}, it is apparent that the real boundaries of the wolf
are well highlighted.
Besides, thanks to the dynamic fusion mechanism, the features learned by DFI
are much more powerful compared to \cite{xie2015holistically,liu2016learning}.
This is because the areas with no edges are rendered much cleaner.
To sum up, in spite of the improvements in ODS and OIS,
the quality of our results is much higher visually.

\begin{table}[t]
  \centering
  \footnotesize
  \renewcommand{\arraystretch}{1.25}
  \renewcommand{\tabcolsep}{0.5mm}
  \caption{Average speed (FPS) comparisons between DFI and the previous state-of-the-art methods.
  } \label{tab:time}
  \begin{tabular}{c|ccccc} \whline{1pt}
          & \textbf{DFI(Multi)}    & \textbf{DFI(Single)} & BASNet \cite{qin2019basnet} & AFNet\cite{feng2019attentive} & PiCANet\cite{liu2018picanet} \\\hline
    Size  & $400\times300$ & $400\times300$ & $256\times256$ & $224\times224$ & $224\times224$  \\\hline
    Speed & 40      &    57    &   25    &   26    &    7    \\
    \whline{1pt}
          & PAGE \cite{wang2019salient} & CPD \cite{wu2019cascaded}  & DGRL \cite{wang2018detect} & Amulet \cite{zhang2017amulet} & DSS \cite{hou2016deeply} \\\hline
    Size  & $224\times224$ & $352\times352$ & $384\times384$ & $256\times256$ & $400\times300$ \\\hline
    Speed &   25    &   61    &   8    &  16      &  12     \\
    \whline{1pt}
    & RCF \cite{RcfEdgePami2019} & CED \cite{wang2017deep}  & LPCB \cite{deng2018learning} & Hi-Fi \cite{zhao2018hifi} & DeepFlux \cite{wang2019deepflux} \\\hline
    Size  & $480\times320$ & $480\times320$ & $480\times320$ & $300\times200$ & $300\times200$ \\\hline
    Speed &   36    &   35    &   35    &  32      &  55     \\
    \whline{1pt}
  \end{tabular}
\end{table}

\subsection{Skeleton Extraction}
We compare DFI with 9 recent CNN-based methods including MIL \cite{tsogkas2012learning}, 
HED \cite{xie2015holistically}, RCF \cite{RcfEdgePami2019}, FSDS \cite{shen2016object}, 
LMSDS \cite{shen2017deepskeleton}, SRN \cite{ke2017srn}, LSN \cite{liu2018linear},
Hi-Fi \cite{zhao2018hifi}, and DeepFlux \cite{wang2019deepflux} on 
2 popular and challenging datasets including SK-LARGE \cite{shen2017deepskeleton} and SYM-PASCAL \cite{ke2017srn}.
For fair comparisons, we train two different models using these two datasets separately,
as done in the above methods.

\myPara{Quantitative Analysis} In \tabref{tab:skel_comps}, 
we show quantitative comparisons with existing methods.
As can be seen, DFI wins dramatically by a large margin (1.9 points) on the SK-LARGE dataset \cite{shen2016object}.
There is also an improvement of 0.9 points on the SYM-PASCAL dataset \cite{ke2017srn}.

\myPara{PR Curves}
In \figref{fig:skel_pr}, we also show the precision-recall curves of our approach with
some selected skeleton extraction methods.
As can be seen, quantitatively, our approach on both datasets 
substantially outperforms other existing methods with a clear margin.

\myPara{Visual Analysis} In \figref{fig:sk_vis}, we show some visual comparisons.
Owing to the advanced features integration strategy that is performed dynamically,
DFI is able to locate
the exact positions of the skeletons more accurately.
This point can also be substantiated by the fact that our prediction maps are much thinner
and stronger than other works.
Both quantitative and visual results unveil that DFI provides a better way
to combine different-level features for skeleton extraction, 
even in a multi-task manner.
%

\subsection{Comparisons of Running Time}
As shown in \tabref{tab:time},
we compare the speed of DFI against other open source 
methods evaluated in our paper including all three tasks.
We report average speed (fps) of different methods 
as well as the corresponding input size below
(tested in the same environment).
DFI can run at 57 FPS in single-task mode which is comparable to other methods while producing better detection results.
Also, DFI runs at 40 FPS even in multi-task mode which means predicting 
three different tasks at the same time.

\section{Conclusion} \label{sec:conclusion}
In this paper, we solve three different low-level pixel-wise prediction tasks simultaneously,
including salient object segmentation, edge detection, and skeleton extraction.
We propose a dynamic feature integration module (DFIM) to learn
the feature integration strategy for each task dynamically,
and a task-adaptive attention module (TAM) to allocate information across tasks for better overall convergence.
Experiments on a wide range of datasets show that DFI can perform 
comparably sometimes even better than the state-of-the-art methods of the solved
tasks.
DFI is fast as well, which can perform these three pixel-wise prediction tasks 
simultaneously with a 
speed of 40 FPS.

\section*{Acknowledgments}
This research was supported by Major Project for New Generation of AI under 
Grant No. 2018AAA0100400, NSFC (61922046), 
the national youth talent support program, 
and Tianjin Natural Science Foundation (18ZXZNGX00110).

\ifCLASSOPTIONcaptionsoff
  \newpage
\fi

\bibliographystyle{IEEEtran}
\bibliography{dfi}

\vspace{-.3in}
\begin{IEEEbiography}[{\includegraphics[width=1in,keepaspectratio]{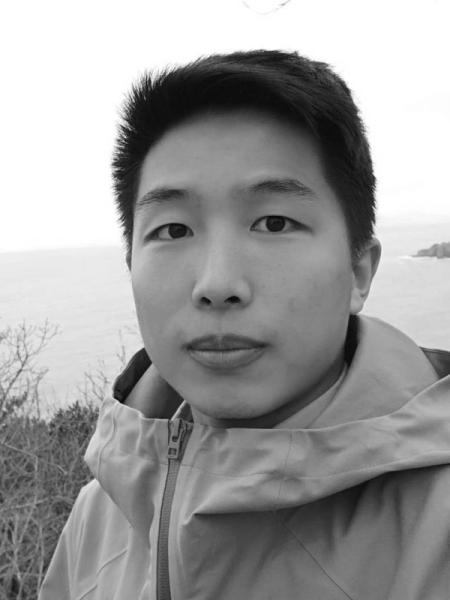}}]
{Jiang-Jiang Liu} is currently a Ph.D. candidate with School of Computer Science,
Nankai University, under the supervision of Prof. Ming-Ming Cheng.
His research interests include deep learning,
image processing, and computer vision.
\end{IEEEbiography}

\vspace{-.4in}
\begin{IEEEbiography}[{\includegraphics[width=1in,keepaspectratio]{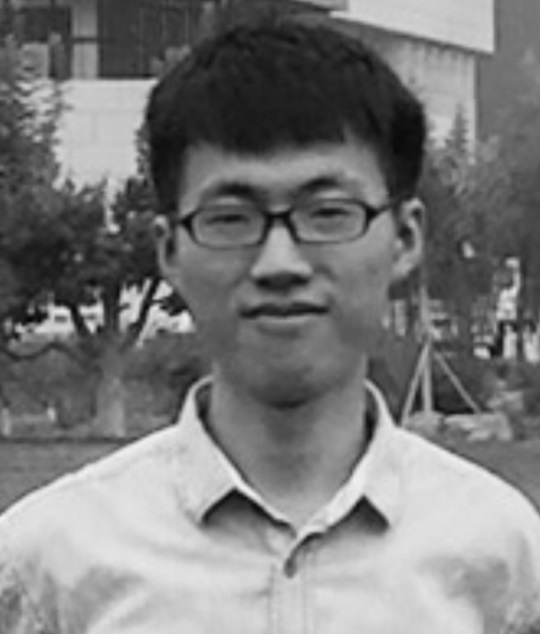}}]
{Qibin Hou} received his PhD degree from Nankai University in 2019.
He is currently a research fellow at the
Department of Electrical and Computer Engineering, National University of Singapore, working with Prof. Jiashi Feng.
His research interests include deep learning and computer vision.
\end{IEEEbiography}

\vspace{-.4in}
\begin{IEEEbiography}[{\includegraphics[width=1in,keepaspectratio]{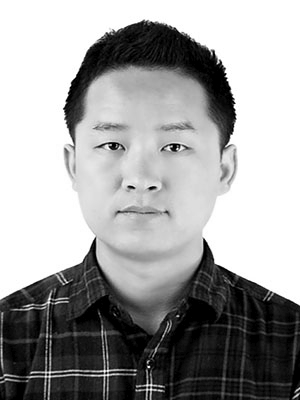}}]
{Ming-Ming Cheng} received his PhD degree from Tsinghua University in 2012.
Then he did 2 years research fellow, with Prof. Philip Torr in Oxford.
He is now a professor at Nankai University, leading the Media Computing Lab.
His research interests includes computer graphics, computer vision, and image processing.
He has published  $60+$ refereed research papers,
with $15,000+$ Google Scholar citations.
He received research awards including ACM China Rising Star Award,
IBM Global SUR Award, \etc
He is a senior member of IEEE and on the editor board of IEEE TIP.
\end{IEEEbiography}

\vfill

\end{document}